\documentclass{article}

\usepackage{microtype}
\usepackage{graphicx}
\usepackage{adjustbox}
\usepackage{pgfplots}
\pgfplotsset{compat=newest}
\usepackage{subcaption}
\usepackage{booktabs} 

\usepackage[accepted]{icml2025}

\usepackage[utf8]{inputenc} 
\usepackage[T1]{fontenc}    
\usepackage{hyperref}       
\usepackage{url}            
\usepackage{booktabs}       
\usepackage{amsfonts}       
\usepackage{nicefrac}       
\usepackage{microtype}      
\usepackage{fancyhdr}       
\usepackage[affil-it]{authblk}

\usepackage{hyperref}

\usepackage{amsfonts,bm}
\usepackage{amsmath}
\usepackage{amssymb}
\usepackage{mathtools}
\usepackage{amsthm}

\usepackage[capitalize,noabbrev]{cleveref}
\usepackage{xcolor}
\usepackage{xspace}
\usepackage{tcolorbox}
\usepackage{pgfplots}
\usepackage{caption}
\usepackage{wrapfig}
\usepackage{color,soul}
\usepackage{natbib}

\theoremstyle{plain}

\theoremstyle{definition}

\theoremstyle{remark}

\usepackage[textsize=tiny]{todonotes}

\begin{document}

\twocolumn[
\icmltitle{\textsc{Teddy}: A FAMILY OF FOUNDATION MODELS FOR
UNDERSTANDING SINGLE CELL BIOLOGY}




\begin{icmlauthorlist}
\icmlauthor{Alexis Chevalier}{yyy}
\icmlauthor{Soumya Ghosh\textsuperscript{+}}{comp}
\icmlauthor{Urvi Awasthi}{yyy}
\icmlauthor{James Watkins}{comp}
\icmlauthor{Julia Bieniewska}{yyy}
\icmlauthor{Nichita Mitrea}{sch}
\icmlauthor{Olga Kotova}{comp}
\icmlauthor{Kirill Shkura}{sch}
\icmlauthor{Andrew Noble}{comp}
\icmlauthor{Michael J. Steinbaugh}{comp}
\icmlauthor{Vijay Sadashivaiah}{comp}
\icmlauthor{George Dasoulas}{comp}
\icmlauthor{Julien Delile}{yyy}
\icmlauthor{Christoph Meier}{yyy}
\icmlauthor{Leonid Zhukov}{yyy}
\icmlauthor{Iya Khalil}{comp}
\icmlauthor{Srayanta Mukherjee\textsuperscript{+}}{yyy}
\icmlauthor{Judith Mueller\textsuperscript{+}}{comp}
\end{icmlauthorlist}

\icmlaffiliation{yyy}{BCG AI Science Institute, Boston, USA}
\icmlaffiliation{comp}{Merck \& Co., Inc., Cambridge, MA, USA}
\icmlaffiliation{sch}{MSD (UK) Limited, London, UK}

\icmlcorrespondingauthor{Soumya Ghosh}{soumya.ghosh@merck.com}
\icmlcorrespondingauthor{Srayanta Mukherjee}{mukherjee.srayanta@bcg.com}
\icmlcorrespondingauthor{Judith Mueller}{judith.mueller@merck.com}

\icmlkeywords{Single cell foundation models, ICML}

\vskip 0.3in
]



\printAffiliationsAndNotice{} 

\newcommand{\figleft}{{\em (Left)}}
\newcommand{\figcenter}{{\em (Center)}}
\newcommand{\figright}{{\em (Right)}}
\newcommand{\figtop}{{\em (Top)}}
\newcommand{\figbottom}{{\em (Bottom)}}
\newcommand{\captiona}{{\em (a)}}
\newcommand{\captionb}{{\em (b)}}
\newcommand{\captionc}{{\em (c)}}
\newcommand{\captiond}{{\em (d)}}

\newcommand{\newterm}[1]{{\bf #1}}

\def\figref#1{figure~\ref{#1}}
\def\Figref#1{Figure~\ref{#1}}
\def\twofigref#1#2{figures \ref{#1} and \ref{#2}}
\def\quadfigref#1#2#3#4{figures \ref{#1}, \ref{#2}, \ref{#3} and \ref{#4}}
\def\secref#1{section~\ref{#1}}
\def\Secref#1{Section~\ref{#1}}
\def\twosecrefs#1#2{sections \ref{#1} and \ref{#2}}
\def\secrefs#1#2#3{sections \ref{#1}, \ref{#2} and \ref{#3}}
\def\eqref#1{equation~\ref{#1}}
\def\Eqref#1{Equation~\ref{#1}}
\def\plaineqref#1{\ref{#1}}
\def\chapref#1{chapter~\ref{#1}}
\def\Chapref#1{Chapter~\ref{#1}}
\def\rangechapref#1#2{chapters\ref{#1}--\ref{#2}}
\def\algref#1{algorithm~\ref{#1}}
\def\Algref#1{Algorithm~\ref{#1}}
\def\twoalgref#1#2{algorithms \ref{#1} and \ref{#2}}
\def\Twoalgref#1#2{Algorithms \ref{#1} and \ref{#2}}
\def\partref#1{part~\ref{#1}}
\def\Partref#1{Part~\ref{#1}}
\def\twopartref#1#2{parts \ref{#1} and \ref{#2}}

\def\ceil#1{\lceil #1 \rceil}
\def\floor#1{\lfloor #1 \rfloor}
\def\1{\bm{1}}
\newcommand{\train}{\mathcal{D}}
\newcommand{\valid}{\mathcal{D_{\mathrm{valid}}}}
\newcommand{\test}{\mathcal{D_{\mathrm{test}}}}
\newcommand{\real}{\mathbb{R}}
\newcommand{\integer}{\mathbb{Z}}
\newcommand{\mlm}{\ell_{\mathrm{MLM}}}
\newcommand{\mlma}{\ell_{\mathrm{MLM-}}}
\newcommand{\mlmg}{\ell_{\mathrm{MLM-G}}}
\newcommand{\mlmx}{\ell_{\mathrm{MLM-X}}}
\newcommand{\annot}{\ell_{\mathrm{CLS}}}
\newcommand{\annotg}{\ell_{\mathrm{CLS-G}}}
\newcommand{\annotx}{\ell_{\mathrm{CLS-X}}}

\def\eps{{\epsilon}}

\def\reta{{\textnormal{$\eta$}}}
\def\ra{{\textnormal{a}}}
\def\rb{{\textnormal{b}}}
\def\rc{{\textnormal{c}}}
\def\rd{{\textnormal{d}}}
\def\re{{\textnormal{e}}}
\def\rf{{\textnormal{f}}}
\def\rg{{\textnormal{g}}}
\def\rh{{\textnormal{h}}}
\def\ri{{\textnormal{i}}}
\def\rj{{\textnormal{j}}}
\def\rk{{\textnormal{k}}}
\def\rl{{\textnormal{l}}}
\def\rn{{\textnormal{n}}}
\def\ro{{\textnormal{o}}}
\def\rp{{\textnormal{p}}}
\def\rq{{\textnormal{q}}}
\def\rr{{\textnormal{r}}}
\def\rs{{\textnormal{s}}}
\def\rt{{\textnormal{t}}}
\def\ru{{\textnormal{u}}}
\def\rv{{\textnormal{v}}}
\def\rw{{\textnormal{w}}}
\def\rx{{\textnormal{x}}}
\def\ry{{\textnormal{y}}}
\def\rz{{\textnormal{z}}}

\def\rvepsilon{{\mathbf{\epsilon}}}
\def\rvtheta{{\mathbf{\theta}}}
\def\rva{{\mathbf{a}}}
\def\rvb{{\mathbf{b}}}
\def\rvc{{\mathbf{c}}}
\def\rvd{{\mathbf{d}}}
\def\rve{{\mathbf{e}}}
\def\rvf{{\mathbf{f}}}
\def\rvg{{\mathbf{g}}}
\def\rvh{{\mathbf{h}}}
\def\rvu{{\mathbf{i}}}
\def\rvj{{\mathbf{j}}}
\def\rvk{{\mathbf{k}}}
\def\rvl{{\mathbf{l}}}
\def\rvm{{\mathbf{m}}}
\def\rvn{{\mathbf{n}}}
\def\rvo{{\mathbf{o}}}
\def\rvp{{\mathbf{p}}}
\def\rvq{{\mathbf{q}}}
\def\rvr{{\mathbf{r}}}
\def\rvs{{\mathbf{s}}}
\def\rvt{{\mathbf{t}}}
\def\rvu{{\mathbf{u}}}
\def\rvv{{\mathbf{v}}}
\def\rvw{{\mathbf{w}}}
\def\rvx{{\mathbf{x}}}
\def\rvy{{\mathbf{y}}}
\def\rvz{{\mathbf{z}}}

\def\erva{{\textnormal{a}}}
\def\ervb{{\textnormal{b}}}
\def\ervc{{\textnormal{c}}}
\def\ervd{{\textnormal{d}}}
\def\erve{{\textnormal{e}}}
\def\ervf{{\textnormal{f}}}
\def\ervg{{\textnormal{g}}}
\def\ervh{{\textnormal{h}}}
\def\ervi{{\textnormal{i}}}
\def\ervj{{\textnormal{j}}}
\def\ervk{{\textnormal{k}}}
\def\ervl{{\textnormal{l}}}
\def\ervm{{\textnormal{m}}}
\def\ervn{{\textnormal{n}}}
\def\ervo{{\textnormal{o}}}
\def\ervp{{\textnormal{p}}}
\def\ervq{{\textnormal{q}}}
\def\ervr{{\textnormal{r}}}
\def\ervs{{\textnormal{s}}}
\def\ervt{{\textnormal{t}}}
\def\ervu{{\textnormal{u}}}
\def\ervv{{\textnormal{v}}}
\def\ervw{{\textnormal{w}}}
\def\ervx{{\textnormal{x}}}
\def\ervy{{\textnormal{y}}}
\def\ervz{{\textnormal{z}}}

\def\rmA{{\mathbf{A}}}
\def\rmB{{\mathbf{B}}}
\def\rmC{{\mathbf{C}}}
\def\rmD{{\mathbf{D}}}
\def\rmE{{\mathbf{E}}}
\def\rmF{{\mathbf{F}}}
\def\rmG{{\mathbf{G}}}
\def\rmH{{\mathbf{H}}}
\def\rmI{{\mathbf{I}}}
\def\rmJ{{\mathbf{J}}}
\def\rmK{{\mathbf{K}}}
\def\rmL{{\mathbf{L}}}
\def\rmM{{\mathbf{M}}}
\def\rmN{{\mathbf{N}}}
\def\rmO{{\mathbf{O}}}
\def\rmP{{\mathbf{P}}}
\def\rmQ{{\mathbf{Q}}}
\def\rmR{{\mathbf{R}}}
\def\rmS{{\mathbf{S}}}
\def\rmT{{\mathbf{T}}}
\def\rmU{{\mathbf{U}}}
\def\rmV{{\mathbf{V}}}
\def\rmW{{\mathbf{W}}}
\def\rmX{{\mathbf{X}}}
\def\rmY{{\mathbf{Y}}}
\def\rmZ{{\mathbf{Z}}}

\def\ermA{{\textnormal{A}}}
\def\ermB{{\textnormal{B}}}
\def\ermC{{\textnormal{C}}}
\def\ermD{{\textnormal{D}}}
\def\ermE{{\textnormal{E}}}
\def\ermF{{\textnormal{F}}}
\def\ermG{{\textnormal{G}}}
\def\ermH{{\textnormal{H}}}
\def\ermI{{\textnormal{I}}}
\def\ermJ{{\textnormal{J}}}
\def\ermK{{\textnormal{K}}}
\def\ermL{{\textnormal{L}}}
\def\ermM{{\textnormal{M}}}
\def\ermN{{\textnormal{N}}}
\def\ermO{{\textnormal{O}}}
\def\ermP{{\textnormal{P}}}
\def\ermQ{{\textnormal{Q}}}
\def\ermR{{\textnormal{R}}}
\def\ermS{{\textnormal{S}}}
\def\ermT{{\textnormal{T}}}
\def\ermU{{\textnormal{U}}}
\def\ermV{{\textnormal{V}}}
\def\ermW{{\textnormal{W}}}
\def\ermX{{\textnormal{X}}}
\def\ermY{{\textnormal{Y}}}
\def\ermZ{{\textnormal{Z}}}

\def\vzero{{\bm{0}}}
\def\vone{{\bm{1}}}
\def\vmu{{\bm{\mu}}}
\def\vtheta{{\bm{\theta}}}
\def\va{{\bm{a}}}
\def\vb{{\bm{b}}}
\def\vc{{\bm{c}}}
\def\vd{{\bm{d}}}
\def\ve{{\bm{e}}}
\def\vf{{\bm{f}}}
\def\vg{{\bm{g}}}
\def\vh{{\bm{h}}}
\def\vi{{\bm{i}}}
\def\vj{{\bm{j}}}
\def\vk{{\bm{k}}}
\def\vl{{\bm{l}}}
\def\vm{{\bm{m}}}
\def\vn{{\bm{n}}}
\def\vo{{\bm{o}}}
\def\vp{{\bm{p}}}
\def\vq{{\bm{q}}}
\def\vr{{\bm{r}}}
\def\vs{{\bm{s}}}
\def\vt{{\bm{t}}}
\def\vu{{\bm{u}}}
\def\vv{{\bm{v}}}
\def\vw{{\bm{w}}}
\def\vx{{\bm{x}}}
\def\vy{{\bm{y}}}
\def\vz{{\bm{z}}}

\def\evalpha{{\alpha}}
\def\evbeta{{\beta}}
\def\evepsilon{{\epsilon}}
\def\evlambda{{\lambda}}
\def\evomega{{\omega}}
\def\evmu{{\mu}}
\def\evpsi{{\psi}}
\def\evsigma{{\sigma}}
\def\evtheta{{\theta}}
\def\eva{{a}}
\def\evb{{b}}
\def\evc{{c}}
\def\evd{{d}}
\def\eve{{e}}
\def\evf{{f}}
\def\evg{{g}}
\def\evh{{h}}
\def\evi{{i}}
\def\evj{{j}}
\def\evk{{k}}
\def\evl{{l}}
\def\evm{{m}}
\def\evn{{n}}
\def\evo{{o}}
\def\evp{{p}}
\def\evq{{q}}
\def\evr{{r}}
\def\evs{{s}}
\def\evt{{t}}
\def\evu{{u}}
\def\evv{{v}}
\def\evw{{w}}
\def\evx{{x}}
\def\evy{{y}}
\def\evz{{z}}

\def\mA{{\bm{A}}}
\def\mB{{\bm{B}}}
\def\mC{{\bm{C}}}
\def\mD{{\bm{D}}}
\def\mE{{\bm{E}}}
\def\mF{{\bm{F}}}
\def\mG{{\bm{G}}}
\def\mH{{\bm{H}}}
\def\mI{{\bm{I}}}
\def\mJ{{\bm{J}}}
\def\mK{{\bm{K}}}
\def\mL{{\bm{L}}}
\def\mM{{\bm{M}}}
\def\mN{{\bm{N}}}
\def\mO{{\bm{O}}}
\def\mP{{\bm{P}}}
\def\mQ{{\bm{Q}}}
\def\mR{{\bm{R}}}
\def\mS{{\bm{S}}}
\def\mT{{\bm{T}}}
\def\mU{{\bm{U}}}
\def\mV{{\bm{V}}}
\def\mW{{\bm{W}}}
\def\mX{{\bm{X}}}
\def\mY{{\bm{Y}}}
\def\mZ{{\bm{Z}}}
\def\mBeta{{\bm{\beta}}}
\def\mPhi{{\bm{\Phi}}}
\def\mLambda{{\bm{\Lambda}}}
\def\mSigma{{\bm{\Sigma}}}

\newcommand{\tens}[1]{\bm{\mathsfit{#1}}}
\def\tA{{\tens{A}}}
\def\tB{{\tens{B}}}
\def\tC{{\tens{C}}}
\def\tD{{\tens{D}}}
\def\tE{{\tens{E}}}
\def\tF{{\tens{F}}}
\def\tG{{\tens{G}}}
\def\tH{{\tens{H}}}
\def\tI{{\tens{I}}}
\def\tJ{{\tens{J}}}
\def\tK{{\tens{K}}}
\def\tL{{\tens{L}}}
\def\tM{{\tens{M}}}
\def\tN{{\tens{N}}}
\def\tO{{\tens{O}}}
\def\tP{{\tens{P}}}
\def\tQ{{\tens{Q}}}
\def\tR{{\tens{R}}}
\def\tS{{\tens{S}}}
\def\tT{{\tens{T}}}
\def\tU{{\tens{U}}}
\def\tV{{\tens{V}}}
\def\tW{{\tens{W}}}
\def\tX{{\tens{X}}}
\def\tY{{\tens{Y}}}
\def\tZ{{\tens{Z}}}

\def\gA{{\mathcal{A}}}
\def\gB{{\mathcal{B}}}
\def\gC{{\mathcal{C}}}
\def\gD{{\mathcal{D}}}
\def\gE{{\mathcal{E}}}
\def\gF{{\mathcal{F}}}
\def\gG{{\mathcal{G}}}
\def\gH{{\mathcal{H}}}
\def\gI{{\mathcal{I}}}
\def\gJ{{\mathcal{J}}}
\def\gK{{\mathcal{K}}}
\def\gL{{\mathcal{L}}}
\def\gM{{\mathcal{M}}}
\def\gN{{\mathcal{N}}}
\def\gO{{\mathcal{O}}}
\def\gP{{\mathcal{P}}}
\def\gQ{{\mathcal{Q}}}
\def\gR{{\mathcal{R}}}
\def\gS{{\mathcal{S}}}
\def\gT{{\mathcal{T}}}
\def\gU{{\mathcal{U}}}
\def\gV{{\mathcal{V}}}
\def\gW{{\mathcal{W}}}
\def\gX{{\mathcal{X}}}
\def\gY{{\mathcal{Y}}}
\def\gZ{{\mathcal{Z}}}

\def\sA{{\mathbb{A}}}
\def\sB{{\mathbb{B}}}
\def\sC{{\mathbb{C}}}
\def\sD{{\mathbb{D}}}
\def\sF{{\mathbb{F}}}
\def\sG{{\mathbb{G}}}
\def\sH{{\mathbb{H}}}
\def\sI{{\mathbb{I}}}
\def\sJ{{\mathbb{J}}}
\def\sK{{\mathbb{K}}}
\def\sL{{\mathbb{L}}}
\def\sM{{\mathbb{M}}}
\def\sN{{\mathbb{N}}}
\def\sO{{\mathbb{O}}}
\def\sP{{\mathbb{P}}}
\def\sQ{{\mathbb{Q}}}
\def\sR{{\mathbb{R}}}
\def\sS{{\mathbb{S}}}
\def\sT{{\mathbb{T}}}
\def\sU{{\mathbb{U}}}
\def\sV{{\mathbb{V}}}
\def\sW{{\mathbb{W}}}
\def\sX{{\mathbb{X}}}
\def\sY{{\mathbb{Y}}}
\def\sZ{{\mathbb{Z}}}

\def\emLambda{{\Lambda}}
\def\emA{{A}}
\def\emB{{B}}
\def\emC{{C}}
\def\emD{{D}}
\def\emE{{E}}
\def\emF{{F}}
\def\emG{{G}}
\def\emH{{H}}
\def\emI{{I}}
\def\emJ{{J}}
\def\emK{{K}}
\def\emL{{L}}
\def\emM{{M}}
\def\emN{{N}}
\def\emO{{O}}
\def\emP{{P}}
\def\emQ{{Q}}
\def\emR{{R}}
\def\emS{{S}}
\def\emT{{T}}
\def\emU{{U}}
\def\emV{{V}}
\def\emW{{W}}
\def\emX{{X}}
\def\emY{{Y}}
\def\emZ{{Z}}
\def\emSigma{{\Sigma}}

\newcommand{\etens}[1]{\mathsfit{#1}}
\def\etLambda{{\etens{\Lambda}}}
\def\etA{{\etens{A}}}
\def\etB{{\etens{B}}}
\def\etC{{\etens{C}}}
\def\etD{{\etens{D}}}
\def\etE{{\etens{E}}}
\def\etF{{\etens{F}}}
\def\etG{{\etens{G}}}
\def\etH{{\etens{H}}}
\def\etI{{\etens{I}}}
\def\etJ{{\etens{J}}}
\def\etK{{\etens{K}}}
\def\etL{{\etens{L}}}
\def\etM{{\etens{M}}}
\def\etN{{\etens{N}}}
\def\etO{{\etens{O}}}
\def\etP{{\etens{P}}}
\def\etQ{{\etens{Q}}}
\def\etR{{\etens{R}}}
\def\etS{{\etens{S}}}
\def\etT{{\etens{T}}}
\def\etU{{\etens{U}}}
\def\etV{{\etens{V}}}
\def\etW{{\etens{W}}}
\def\etX{{\etens{X}}}
\def\etY{{\etens{Y}}}
\def\etZ{{\etens{Z}}}

\newcommand{\pdata}{p_{\rm{data}}}
\newcommand{\ptrain}{\hat{p}_{\rm{data}}}
\newcommand{\Ptrain}{\hat{P}_{\rm{data}}}
\newcommand{\pmodel}{p_{\rm{model}}}
\newcommand{\Pmodel}{P_{\rm{model}}}
\newcommand{\ptildemodel}{\tilde{p}_{\rm{model}}}
\newcommand{\pencode}{p_{\rm{encoder}}}
\newcommand{\pdecode}{p_{\rm{decoder}}}
\newcommand{\precons}{p_{\rm{reconstruct}}}

\newcommand{\laplace}{\mathrm{Laplace}} 

\newcommand{\E}{\mathbb{E}}
\newcommand{\Ls}{\mathcal{L}}
\newcommand{\R}{\mathbb{R}}
\newcommand{\emp}{\tilde{p}}
\newcommand{\lr}{\alpha}
\newcommand{\reg}{\lambda}
\newcommand{\rect}{\mathrm{rectifier}}
\newcommand{\softmax}{\mathrm{softmax}}
\newcommand{\sigmoid}{\sigma}
\newcommand{\softplus}{\zeta}
\newcommand{\KL}{D_{\mathrm{KL}}}
\newcommand{\Var}{\mathrm{Var}}
\newcommand{\standarderror}{\mathrm{SE}}
\newcommand{\Cov}{\mathrm{Cov}}
\newcommand{\normlzero}{L^0}
\newcommand{\normlone}{L^1}
\newcommand{\normltwo}{L^2}
\newcommand{\normlp}{L^p}
\newcommand{\normmax}{L^\infty}
\newcommand{\cxg}{\textsc{CELLxGENE }}
\newcommand{\annotdis}{\vy^{\texttt{<disease>}}_{{n}}}
\newcommand{\annotcell}{\vy^{\texttt{<cell>}}_{{n}}}
\newcommand{\annottissue}{\vy^{\texttt{<tissue>}}_{{n}}}
\newcommand{\annotsex}{\vy^{\texttt{<sex>}}_{{n}}}
\newcommand{\preddisgene}{g^{\text{dcls}}_\theta}
\newcommand{\predcellgene}{g^{\text{ccls}}_\theta}
\newcommand{\predtissuegene}{g^{\text{tcls}}_\theta}
\newcommand{\predsexgene}{g^{\text{scls}}_\theta}
\newcommand{\nicheformer}{\textsc{Nicheformer}}
\newcommand{\geneformer}{\textsc{Geneformer}}
\newcommand{\scGPT}{\textsc{scGPT}}
\newcommand{\model}{\textsc{Teddy}\xspace}
\newcommand{\modelgene}{\textsc{Teddy-G}\xspace}
\newcommand{\modelrank}{\textsc{Teddy-X}\xspace}

\newcommand{\parents}{Pa} 

\let\ab\allowbreak
\begin{abstract}
Understanding the biological mechanisms of disease is crucial for medicine, and in particular, for drug discovery. AI-powered analysis of genome-scale biological data holds great potential in this regard. The increasing availability of single-cell RNA sequencing data has enabled the development of large foundation models for disease biology. However, existing foundation models only modestly improve over task-specific models in downstream applications. Here,  we explored two avenues for improving single-cell foundation models. First, we scaled the pre-training data to a diverse collection of 116 million cells, which is larger than those used by previous models. Second, we leveraged the availability of large-scale biological annotations as a form of supervision during pre-training. We trained the \model family of models comprising six transformer-based state-of-the-art single-cell foundation models with 70 million, 160 million, and 400 million parameters. We vetted our models on several downstream evaluation tasks, including identifying the underlying disease state of held-out donors not seen during training, distinguishing between diseased and healthy cells for disease conditions and donors not seen during training, and probing the learned representations for known biology. Our models showed substantial improvement over existing works, and scaling experiments showed that performance improved predictably with both data volume and parameter count.  
\end{abstract}

\section{Introduction}
\label{sec:intro}
The complexity of cell biology and the mechanisms of disease pathogenesis are driven by an intricate regulatory network of genes \citep{annurev:/content/journals/10.1146/annurev-genom-091416-035537,theodoris2015human,theodoris2021network}. A better resolution of this complex interactome network would enhance our ability to design drugs that target the causal mechanism of the disease rather than interventions that aim to modulate the downstream effects~\citep{ding2022temporal}. However, accurate inference of gene regulatory networks is challenging. The possible space for genetic interactions is vast~\citep{bunne2024build}, the networks to be inferred are highly context-dependent, different cell types and tissue types exhibit different regulatory networks and exhibit significant variations across donors~\citep{chen2024robust}. Moreover, the data required to study gene regulatory networks for a specific disease is usually limited and highly specialized, often plagued by experimental artifacts~\citep{hicks2018missing}. 

However, a confluence of recent technological progress promises to make this challenging problem more tractable. The advent of accurate single-cell sequencing technologies that remove the artifacts of bulk cell data, better reflect natural variability, and provide signals at higher resolutions. This, along with the increasing availability of atlas-scale scRNA-seq datasets that span an extensive range of diseases, cell types, tissue types, and donors provide an unprecedented opportunity for studying disease mechanisms at scale. Parallel and complementary progress~\citep{vaswani2017attention} in artificial intelligence has produced tools, so-called foundation models, capable of absorbing and effectively learning from massive amounts of data~\citep{devlin2018bert,radford2019language,brown2020languagemodelsfewshotlearners, grattafiori2024llama3herdmodels, achiam2023gpt}. These approaches use self-supervised learning to learn from large volumes of unlabeled data, for instance, data scraped from all of the internet, and then adapt to specific tasks from modest amounts of task-specific, typically labeled, data~\citep{devlin2018bert, brown2020languagemodelsfewshotlearners} and have revolutionized the ability of algorithms to understand natural language and images. Whether the same learning paradigm can be used to enhance our understanding of the language of genes, specifically disease biology, has been a subject of recent research~\citep{theodoris2023transfer, cui2024scgpt, schaar2024nicheformer, Chen2024}. This interest is fueled by the fact that seemingly many of the basic principles that drive success in natural language carry over to the gene space: gene embeddings, i.e., gene representations in the vector space, can capture rich context-dependent relations \citep{wagner2016revealing} that can be learned at scale from single-cell measurements in an unsupervised fashion.

Despite recent progress, open questions remain about the usefulness and the potential of foundation models for scRNA-seq. 
\citeauthor{theodoris2023transfer} showed that downstream performance correlates with pre-training dataset size, but others~\citep{liu2023evaluating, boiarsky2023deep} found that untrained transformer models and traditional machine learning methods are competitive and can sometimes surpass foundation models pre-trained on large single cell corpora. Yet others~\citep{kedzierska2023assessing} found the zero-shot embeddings extracted from various foundation models to not substantially improve over less resource-intensive approaches. Moreover, methods for extracting gene regulatory information from foundation models remain under explored, with existing attempts~\citep{cui2024scgpt} relying on simple clustering methods over gene embedding tables. 

In this work, we took a systematic approach to the development of single-cell foundation models, seeking to advance the state-of-the-art in understanding disease biology. We trained a new family of foundation models adapted to disease biology, the \model family of models\footnote{\model: \emph{Transformer for Enabling Drug DiscoverY}}. 
These models incorporate architectural choices inspired by existing works \scGPT~\cite{cui2024scgpt}, \geneformer~\citep{theodoris2023transfer}, and \nicheformer~\citep{schaar2024nicheformer}, train on to-date the largest corpus of single-cell data, and use biological ontologies to supervise gene and cell representation learning. \model is trained on data sourced from \cxg~\citep{CellXGene}. The pre-training dataset contains 116 million (116M) cells from mouse, human, spatial, and dissociated scRNA-seq data. The \model family ranges from 10M to 400M parameters, allowing us to examine the scaling behavior with respect to the number of parameters and the amount of pre-training data. To evaluate performance on the downstream task of disease understanding, we benchmark \model against existing models on two disease classification tasks. We find that the larger \model models improve on existing models on one task substantially and are within noise of the best-performing competitor in the other. 

We believe \model is an important step towards designing foundation models that understand disease biology. Nonetheless, significant innovations such as principled approaches for incorporating existing biological knowledge, continuing to scale the amount of pre-training data, and incorporating other complementary modalities will all likely be required to design foundation models that fully represent the complexity of single-cell biology.

\section{Related work}
\label{sec:related}

The success of transformer-based foundation models in modeling natural language~\citep{devlin2018bert, achiam2023gpt} and vision~\citep{radford2021learningtransferablevisualmodels} has inspired a growing body of research in modeling single-cell transcriptomic data similarly. These models, analogously to natural language, treat cells as sentences and genes expressed in cells as words. They primarily differ in how they represent gene expression data, the training data quality and volume, and the self-supervised objectives they use for learning. We highlight these design choices below. 

Owing to the lack of natural ordering among genes expressed in a cell, BERT~\citep{devlin2018bert} style models that employ bi-directional self-attention and are trained via masked language modeling (MLM) are popular in the literature. scBERT~\citep{yang2022scbert}, an early example of this category, was trained on a corpus of one million (1M) cells from PanglaoDB~\citep{franzen2019panglaodb} and primarily evaluated on the downstream tasks of cell-type annotation and discovery.  

Subsequent models, including the \geneformer v1~\citep{theodoris2023transfer} and v2~\citep{Chen2024}, were trained on larger corpora of 30M and 95M cells. These works represent a cell as a ranked list of genes it expresses. This rank-value encoding scheme ranks genes expressed in a cell based on their (normalized) expression values from most to least expressed. The pre-training task then involves predicting the gene expressed at a particular position in the ranked list, having observed the other genes in the ranked list. The authors demonstrate that pre-training in this fashion endows the models with diverse downstream capabilities, including network and chromatin dynamic predictions, in-silico gene-network analysis, and batch integration. Another recent model, \nicheformer~\citep{schaar2024nicheformer}, also makes use of rank-value encoding but trains on a corpus of 110M cells containing both dissociated (53M) and spatially resolved (57M) cells drawn from both humans and mice to decode spatially resolved cellular information.

Others, for instance, \scGPT \citep{cui2024scgpt}, a 53M\footnote{\href{https://virtualcellmodels.cziscience.com/model/0193323f-2875-7858-862c-6903bf667543?utm_source=czi&utm_campaign=MVP_launch&utm_medium=blog}{CZI blog}} parameter model trained on 33M cells, bin gene expression counts independently for each cell and train the model to predict the bin a gene's expression in a cell discretizes to given the other binned gene expressions and optional meta-data about the cell. Yet others, scFoundation~\citep{hao2024large}, AIDO.Cell~\citep{Ho2024} represent gene expressions as weighted combinations of embeddings, where the mapping from the expression level of a gene to weights in the weighted combination are learned along with the embeddings during pre-training. They are trained on a corpus of 50M cells on a pre-training task of predicting observed gene expression levels from corrupted versions. Decoder-only architectures~\citep{bian2024scmulan} and approaches~\citep{rosen2023universal} combining transcriptomic foundation models with foundation models for other modalities are also starting to emerge. 

Beyond model development, there is also fledgling literature focused on evaluating the effectiveness of foundation models~\citep{liu2023evaluating, kedzierska2023assessing, boiarsky2023deep} on various downstream tasks and reporting mixed results with foundation models showing promise but not consistently outperforming task-specific traditional methods that do not leverage large cell atlases.

Motivated by the shortcomings of current approaches, we explore two avenues for improving over the state-of-the-art. First, we scale the pre-training dataset to 116M cells which is an order of magnitude larger than data scBert, scMulan, and scGPT~\citep{yang2022scbert, bian2024scmulan} were pre-trained on and a few million cells larger than those used by Nicheformer~\citep{schaar2024nicheformer} and Geneformerv2~\citep{Chen2024}. Second, we note that in contrast with domains like text and natural images, where training datasets are largely scraped from the internet, experimental data in scRNA-seq atlases such as \cxg come with rich meta-data annotations that can aid in learning better representations. We explicitly leverage these annotations while training our \model family of models by augmenting the self-supervised pre-training task of masked language modeling with an additional supervised task of predicting available meta-data annotations from gene expressions.

\section{The \model family of foundation models}
\label{sec:Teddy}
The \model family of models contains two variants that differ in how they represent gene expressions and what pre-training tasks they use. \modelgene follows the $\geneformer$ recipe. It uses rank-value encoding and trains on the self-supervised task of 
predicting the gene expressed at a particular position in
the ranked list representing a cell. \modelrank follows the $\scGPT$ approach of binning gene expressions and training the model to predict the binned expression level of a gene. Both \modelgene and \modelrank additionally use the supervised annotation loss detailed in \cref{sec:bioannot}. We train a series of performant models for both variants containing approximately 70M, 160M, and 400M parameters. We also train smaller, less effective 10M and 30M parameter models to carefully probe the scaling behavior of these models and find that the two approaches present different challenges for downstream applications.

\subsection{Pre-training dataset}
We derive our training corpus from \cxg~\citep{CellXGene}, a collection of 1,399 single-cell RNA-seq datasets from open-source publications. At the time of download\footnote{June 11, 2024.}, \cxg contained 160M cells, of which 70M were from primary datasets, from 24,000 donors, 122 different diseases, 413 tissue types, 860 cell types, and 23M diseased cells. 

We filtered low quality cells containing fewer than 225 gene counts and more than 10\% mitochondrial transcript abundance. Dying or highly stressed cells often exhibit high levels of mitochondrial gene expression and were removed following recommendations from previous work~\citep{Satija2023}.  We also excluded studies using 10x Genomics Chromium v1 chemistry in favor of newer datasets captured with v2 and v3 chemistry. After quality control, our corpus contained 116M cells. We held out a subset of the data for validation and removed all data sets used for downstream testing in \cref{sec:benchmark} to avoid data contamination. 

\subsection{Supervision via biological annotations}
\label{sec:bioannot}
\begin{table}[ht]
\centering
\caption{\textbf{Biological annotations.} Each ontology term in \cxg is mapped to one of the forty-three categories below, and the corresponding special token is added to the vocabulary.}
\resizebox{1\linewidth}{!}{
\begin{tabular}{l p{11cm}}
\toprule
\emph{Special tokens} & \emph{Annotation labels} \\
\midrule
\texttt{<disease>} & brain, cancer, cardiovascular, genetic, immune, infectious, kidney, respiratory, other, healthy \\
\texttt{<tissue\_type>}& adipose, cardiovascular, central nervous, digestive, embryonic, endocrine, exocrine, eye, hematopoietic, hepatic, integumentary, musculature, renal, reproductive, respiratory, sensory, unknown \\
\texttt{<cell\_type>} & ciliated, connective, contractile, embryonic, epithelial, hematopoietic, immune, neural, perivascular, precursor, secretory, skeletal, unknown \\
\texttt{<sex>} & male, female, unknown \\
\bottomrule
\end{tabular}
}
\label{tab:ontology_mapping}
\end{table}
Annotations such as disease, cell type, tissue type, sex, and other labels capturing cell-specific metadata are available in \cxg data. Existing foundation models either do not use these annotations during pre-training or only use these labels as metadata tokens to provide additional context for gene modeling, i.e., predicting gene expression levels in a cell. In early experiments, we found that adding metadata as context for gene modeling brought no measurable improvements in gene modeling abilities (data not shown). Instead, we leverage the available annotations as supervisory signals during pre-training. Our pre-training task involves biological annotation modeling, i.e., predicting biological annotations associated with a cell, in addition to gene modeling. This encourages models to learn embeddings that, in addition to being predictive of gene expression levels, align with high-level biological properties encoded in the annotations. 

Focusing on disease, cell type, tissue type, and sex annotations, we use the \cxg ontology tree to organize the $1399$ labels that occur in these four categories into forty-three broad labels, which correspond to metadata labels for the data sets in the \cxg corpus, shown in \cref{tab:ontology_mapping}. 
We emphasize that the ontology terms we train with are coarse-grained annotations. They do not include fine-grained disease labels used in downstream model evaluation. 

To incorporate the biological annotations, we add four special tokens \texttt{<disease>}, \texttt{<tissue\_type>}, \texttt{<cell\_type>}, \texttt{<sex>} to the model's vocabulary, one for each annotation category of interest. We prepend these special tokens to the sequence of genes expressed by a cell and train the model to predict the annotation labels from gene expression data. 
\subsection{Pre-training objective} \label{objective}
\begin{figure}
\includegraphics[width=1\linewidth]{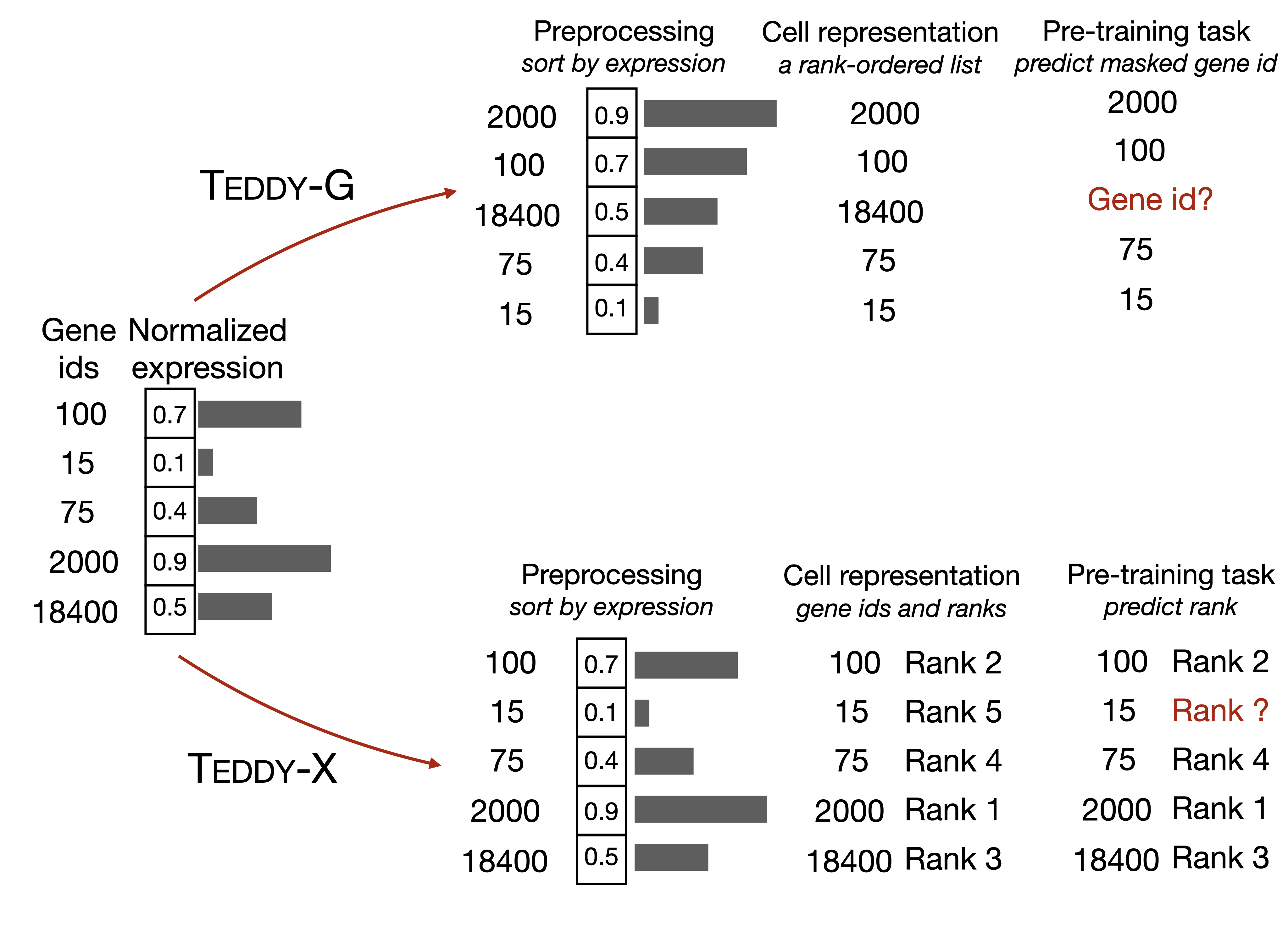}
\caption{\textbf{An illustration of differences between \model variants}. On the left we illustrate a cell with five non-zero expressed genes with non-zero median normalized expression values. \modelgene represents a cell as a list of genes ordered by their expression levels and the pre-training task involves predicting the index of masked out genes. \modelrank ranks expression values, scales them to the interval $[-1, +1]$, and then learns to predict a masked rank scaled to the interval $[-1, +1]$.}
\label{fig:pretrain}
\end{figure}
We introduce notation to state the pre-training objective used by \model variants formally. Let $V$ denote 
the vocabulary size of the model, $\vx_v \in \real^d$ denote a d-dimensional embedding, and $v \in [1, \ldots, V]$ denote the index of gene $v$ in the vocabulary. Denote by $\vx_{\texttt{<disease>}}$, $\vx_{\texttt{<tissue>}}$, $\vx_{\texttt{<cell>}}$, $\vx_{\texttt{<sex>}}$ the embeddings associated with the special tokens. Let $\vt_n =\{t_{nj}\}_{j=1}^{J_n}$ with $t_{nj} \in [1, \ldots, V]$ and  $J_n < V$, represent the indices of the genes expressed by the cell $\vc_n$, the self supervised objective employed by \modelgene is,
$$
    \mlmg(\theta) = \E_{\rvc_n\sim\gD}\big[\E_{m\sim\gM}[-\log \text{Cat}(t_{nm} \vert g_\theta(\rvc_{n\backslash m}))]\big], 
$$
where $\gD = \{\rvc_n\}_{n=1}^\mathrm{N}$ is a dataset containing $\mathrm{N}$ cells, $\theta$ are \modelgene parameters, $\text{Cat}$ represents  a $V$-way categorical distribution, $\gM$ is the set of masked genes, and $\rvc_{n\backslash m} =\{\vx_{nj}\}_{j\neq m} \bigcup \{\vx_{\texttt{<disease>}}, \vx_{\texttt{<tissue>}}, \vx_{\texttt{<cell>}}, \vx_{\texttt{<sex>}} \}$ represents the unmasked portion of the cell and $g_\theta(\rvc_{n\backslash m})$ is the softmax-transformed mean of the categorical distribution predicted by \modelgene. Throughout this work, we create $\gM$ by masking out $15\%$ of the genes uniformly at random. Let $r_{nj}$ represent the rank of gene $j$ in cell $\vc_n$ scaled to the interval $[-1, 1]$. That is, the most expressed gene in cell $\vc_n$ is mapped to $1$ and the gene with the smallest non-zero expression is mapped to $-1$. The self-supervised objective employed by \modelrank is,
\begin{equation*}
\begin{split}
    &\mlmx(\theta) = \E_{\rvc_n\sim\gD}\big[\E_{m\sim\gM}[-\log \mathcal{N}(r_{nm} \vert f_\theta(\rvc_n), 1)]\big], 
\end{split}
\end{equation*}
where $f_\theta(\rvc_{n})$ is predicted by \modelrank and  ${\rvc_n = \rvc_{n\backslash m} \bigcup \{\vx_{nm}\}}$. 

Let $\annotdis, \annotcell, \annottissue, \annotsex$ denote one-hot vectors representing the disease, cell-type, tissue-type, and sex annotations associated with cell $\vc_n$. The supervised component of the pre-trained objective is,
\begin{equation*}
\begin{split}
    &\annotg(\theta) \\&= \E_{\rvc_n\sim\gD}\big[\E_{m\sim\gM}[-\log \text{Cat}(\annotdis \vert \preddisgene(\rvc_{n\backslash m})) \\
    & - \log \text{Cat}(\annotcell \vert \predcellgene(\rvc_{n\backslash m})) \\
    &- \log \text{Cat}(\annottissue \vert \predtissuegene\rvc_{n\backslash m})) \\
    & - \log \text{Cat}(\annotsex \vert \predsexgene(\rvc_{n\backslash m}))
    ]\big], 
\end{split}
\end{equation*}
where $\preddisgene(\rvc_{n\backslash m})), \predcellgene(\rvc_{n\backslash m})), \predtissuegene(\rvc_{n\backslash m})), \predsexgene(\rvc_{n\backslash m}))$ represent \modelgene's softmax-transformed predictions. $\annotx$ is analogously defined. Finally, the overall pre-training objective for \modelgene is, 
\begin{equation}
\ell_{\mathrm{pre-G}}(\theta) = \mlmg(\theta) + \annotg(\theta).
\label{eq:teddyg-pre}
\end{equation}

The pre-training objective for \modelrank involves replacing $\mlmg$ with $\mlmx$ and $\annotg$ with $\annotx$ in \cref{eq:teddyg-pre}. In preliminary experiments, we found weighting the two components of the pre-training objective differently did not substantially affect pre-training or downstream performance. A more careful exploration is part of planned future work. Other recent work~\cite{wang2025limitations} cautions against using granular cell-type labels for supervision, demonstrating that models trained with such supervision distort underlying biological relationships among cells. They instead advocate for weaker supervision. The coarse cell-type labels we employ, along with the additional regularization provided by the masked language modeling loss, provide precisely such weak supervision.
\subsection{Training details and scaling laws}
We used a context length of $2048$ genes for \model models. We based this choice on results from pilot experiments on a small subset of the data (see \cref{subsec:other_design}. We trained all model variants for a single epoch, on 116M cells. We performed non-zero median normalization~\citep{theodoris2021network} of the expression data and use a vocabulary of $48,308$, which includes human and mouse protein-coding and micro-RNA genes as well as special tokens and tokens corresponding to biological annotation labels.  We included micro-RNA genes~\citep{theodoris2023transfer} and mouse data~\citep{schaar2024nicheformer} based on similar choices made in previous work. We used a batch size of $256$, a linear warmup with a linear decay learning rate scheduler with a warmup of $10000$ steps, a maximum learning rate of 1e-4 decayed to zero at the end of training, and the AdamW optimizer. We also found that a relatively large weight decay of $0.1$ was necessary to avoid loss spikes and divergences during training. See \cref{{sec:details}} for additional details and \cref{fig:pretrain} for a summary.

\begin{figure*}[ht]
    \begin{center}
        \includegraphics[width=0.49\linewidth]{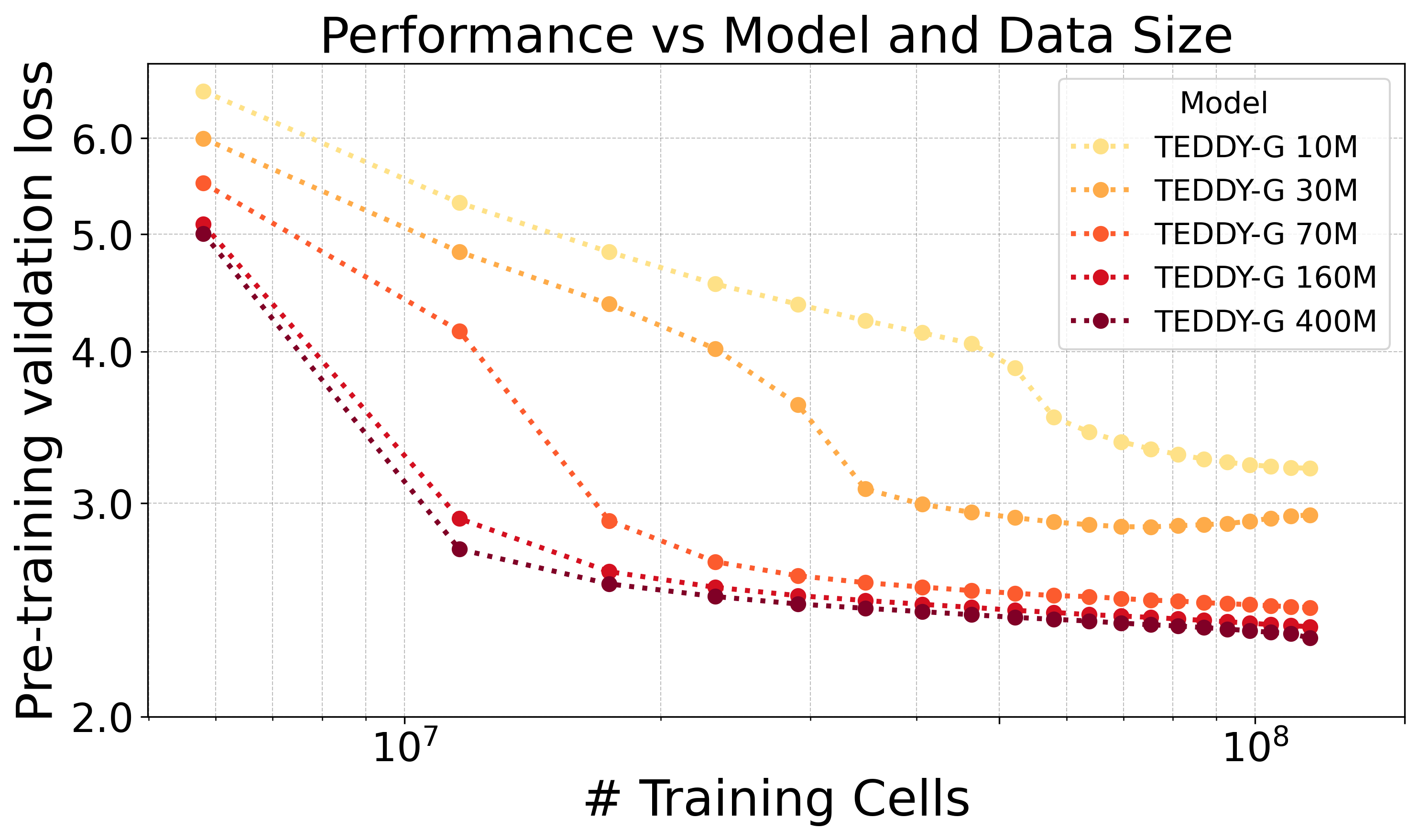}
        \includegraphics[width=0.49\linewidth]{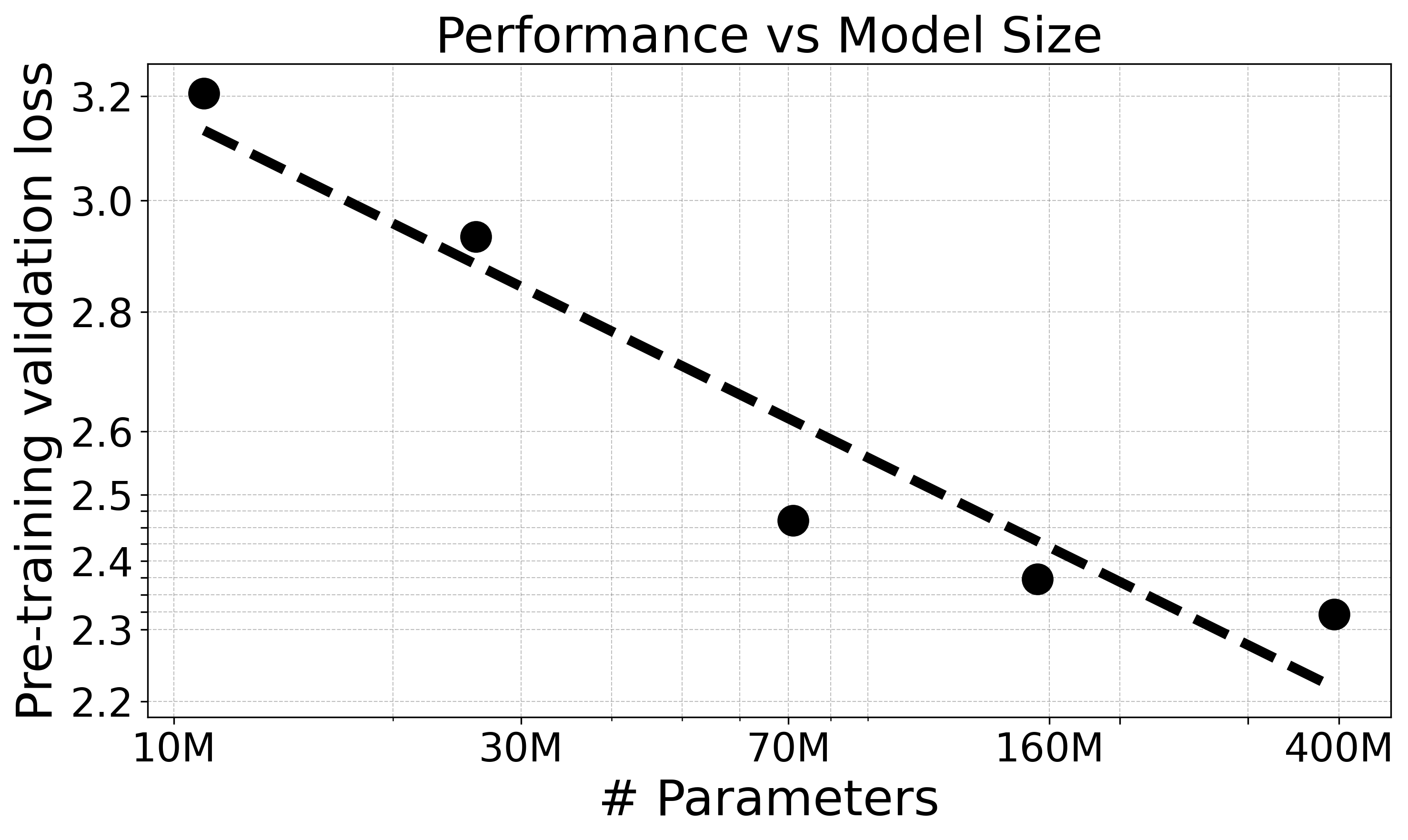}
    \end{center}
    \caption{\textbf{Scaling behavior.} \emph{Left:} Pre-training loss on held-out data for \modelgene as a function of pre-training data and model size. \emph{Right:} Validation loss at the end of one epoch of training against the number of parameters in the model. The dashed line in black is the linear best-fit:  $14.95\times(\text{\# parameters})^{-0.10}$. }
    \label{fig:scalinglaws}
\end{figure*}

\cref{fig:scalinglaws} illustrates performance on held-out data for \modelgene models as a function of model size and training data volume. We observe that performance improves with both increasing model size and data volume but improves only modestly between the 160M and 400M parameter models and not in the straight power law regime.

\section{A disease benchmark for transcriptomic foundation models}
\label{sec:benchmark}
To assess the performance of the TEDDY foundation models, we carried out a benchmarking exercise. The extensive pre-training datasets used to train these models increase the risk of unintentionally contaminating the benchmark evaluation data with information that the model may have encountered during pre-training. Cognizant of these issues, we designed two downstream tasks that hold out data at different levels of granularity to assess the disease modeling capabilities of transcriptomic foundation models.

\subsection{Held-out diseases task} 
In the first task, we evaluate the models' ability to generalize to unseen disease conditions. This involves assessing the models using data from donors with diseases that were not included in the pre-training phase. 

We sourced five datasets from the \cxg database, each containing cells from healthy donors and donors with a single disease condition. We sub-sampled these datasets to ensure that the proportion of diseased and healthy cells is equal. The disease conditions included are \texttt{Chronic Kidney Disease (CKD), Alzheimer's Disease, Gastric Cancer, Rheumatoid Arthritis}, and \texttt{Pediatric Crohn's Disease}. To prevent data contamination, we excluded these five datasets, as well as any other datasets from \cxg that contained these disease labels, from our pre-training corpus. Additional details about the data can be found in \cref{app:evals}.

The benchmark task involves solving five binary classification problems--one for each disease condition--to identify whether a cell is healthy or diseased. We performed three-fold cross validation for these tasks and ensure that there is no donor overlap across folds. The final reported accuracy is averaged over the each of the three held-out test sets in the folds.

\subsection{Held-out donors task} 
We designed the second task to investigate whether foundation models generalize across the biological variability exhibited by different cell donors. To this end, we held out data from 82 donors, with thirteen different disease (or Normal) conditions --- \texttt{COVID-19, Alzheimer's Disease, Acute Myeloid Leukemia, Blastoma, Luminal A Breast Carcinoma, Gingivitis, Luminal B Breast Carcinoma, Multiple Sclerosis, Myocardial Infarction, Normal, Periodontitis, Pilocytic Astrocytoma, Premalignant Hematological System Disease,} and \texttt{type-2 Diabetes Mellitus} from the pre-training corpus.  We then evaluated the effectiveness of different methods at labeling whether a cell comes from a donor with one of these thirteen disease conditions or a healthy donor given the genes expressed by the cell. Unlike the coarse-grained disease annotations during pre-training (\cref{tab:ontology_mapping}), these disease annotations are finer-grained, making this a non-trivial classification problem.
We additionally collected 511 donors from the pre-training corpus  and thirteen donors with Alzhiemer's disease which had been left out of the pre-training dataset for a total of 524 donors for training and validation. We finetuned on 70\% of the 524 donors and held-out out 30\% of the donors for validation. Finally, we tested on the 82 donors that were held out from the pretraining corpus. By construction our train, validation, and test splits had no overlapping donors.  See \cref{app:evals} for additional details about the dataset. 

\section{Vetting the \model family of foundation models}
\label{sec:exp}
With the benchmark in hand, we performed careful experiments to vet the various modeling choices made by \model models. We then proceed to compare \model against existing state-of-the-art foundation models -- 
\nicheformer~\citep{schaar2024nicheformer}, \scGPT~\citep{cui2024scgpt}, six-layer $\geneformer$, and twelve-layer  $\geneformer$ 12L variants~\citep{theodoris2023transfer},  and task-specific machine learning approaches. 

The held-out donors task exhibits class imbalance; we thus report both accuracy and weighted $\text{F}_1$ metrics. The results are aggregated over three random initializations of the foundation models. In the held-out diseases tasks, we evaluate on class-balanced datasets and hence only report accuracies. The results for this task are aggregated over the three cross-validation folds, each using a different random initialization.   

\begin{table*}[t]
    \centering
    \caption{Performance of different foundation models on held-out donors task}
    \resizebox{0.8\textwidth}{!}{
    \begin{tabular}{@{}lccccc@{}}
        \toprule
        & \textbf{\model-G 400M} &\textbf{\nicheformer} & \textbf{\scGPT} & \textbf{\geneformer} & \textbf{\geneformer 12L} \\
        \midrule
        \multicolumn{5}{l}{\textbf{Held-out donors - Fine-tuning}} \\
        \midrule
        \textbf{Accuracy} & 0.72$\pm 0.04$ &0.64$\pm 0.01$ & 0.64$\pm 0.01$ & 0.39$\pm 0.00$ & 0.55$\pm 0.002$ \\
        \textbf{Weighted F1} & 0.68$\pm 0.06$ & 0.56$\pm 0.01$ & 0.54$\pm 0.01$ & 0.22$\pm 0.02$ & 0.45$\pm 0.02$ \\
        \bottomrule
    \end{tabular}}
    \label{tab:donors_evaluation_results}
\end{table*}

\subsection{Downstream performance improves with model scale and biological annotations}
\begin{figure}[t]
    \begin{center}
        \includegraphics[width=0.52\linewidth]{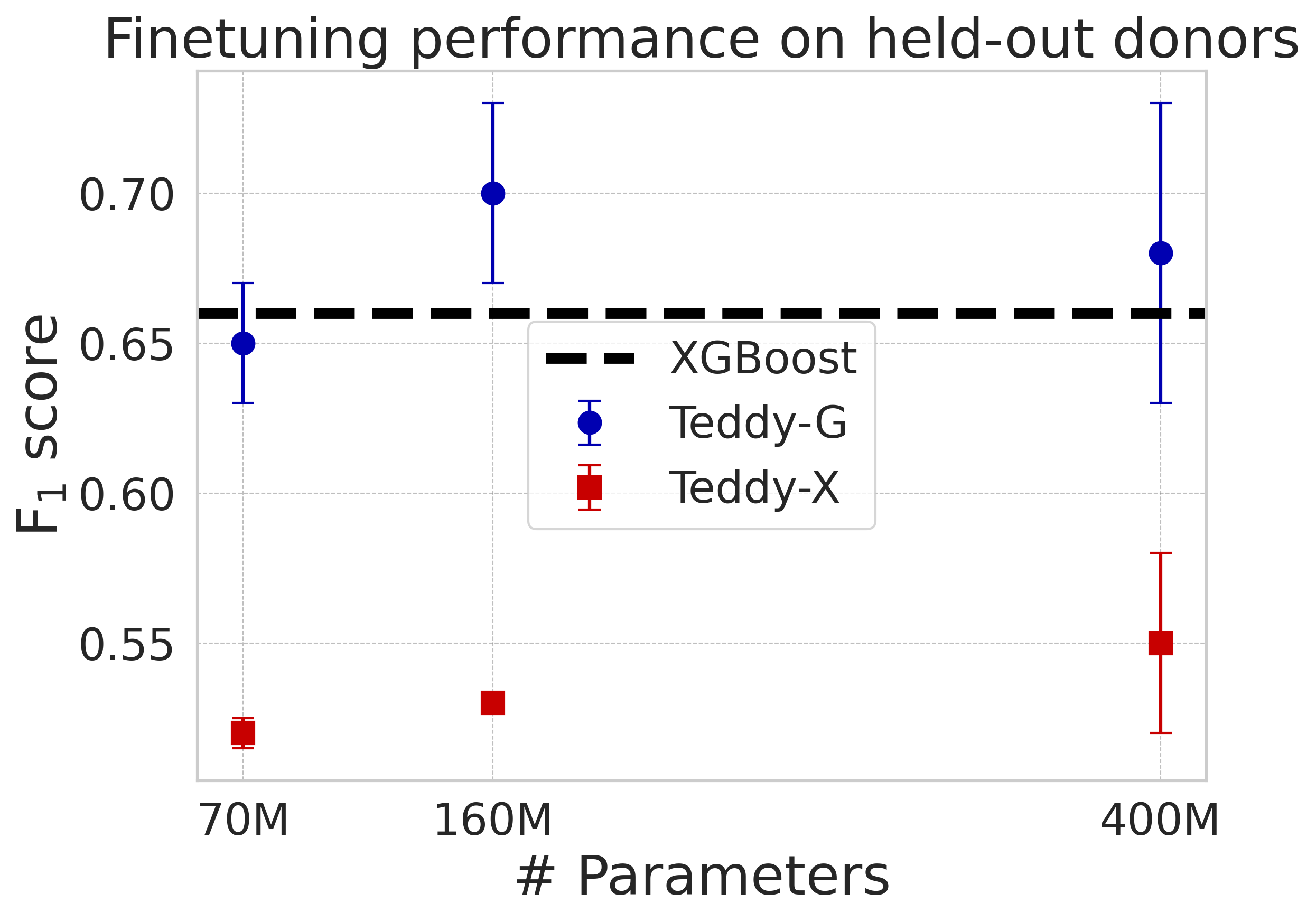}
        \includegraphics[width=0.4\linewidth]{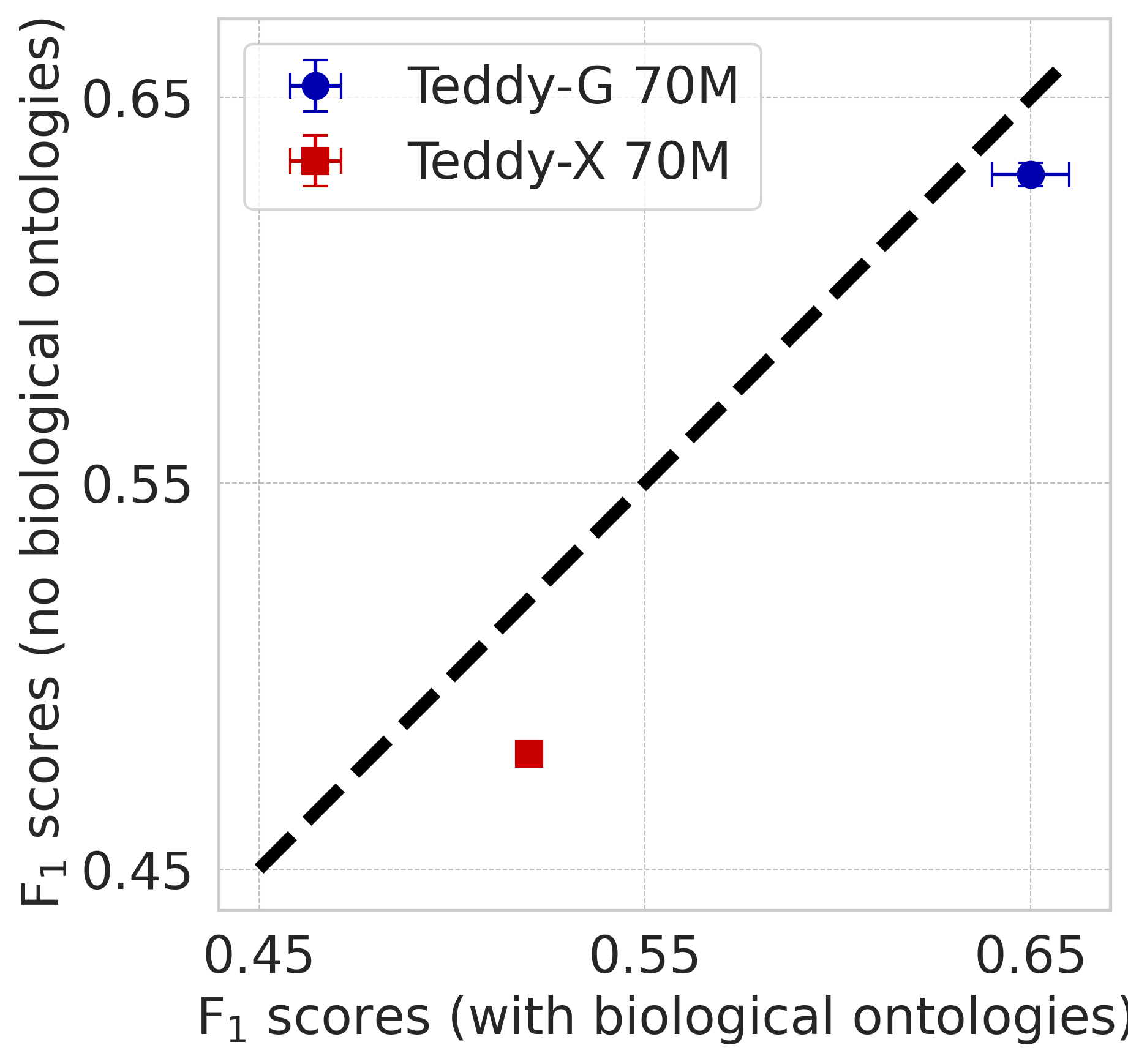}
    \end{center}
    \caption{\textbf{Performance on held-out donors as a function of model size and biological annotaions} \emph{Left:}$\text{F}_1$ scores improve for both \modelgene and \modelrank with increasing size. \modelgene outperforms \modelrank across model scale and improves on XGBoost trained exclusively for this task. \emph{Right:} The x-axis plots the $\text{F}_1$ scores achieved by \modelgene 70M and \modelrank 70M when pre-trained with supervision from biological ontologies. The y-axis plots $\text{F}_1$ scores achieved by the same models without ontologies.}
    \label{fig:downstream_and_ontologies}
\end{figure}

We begin with experiments comparing the impact of model size on our evaluations benchmark. \cref{fig:downstream_and_ontologies} plots fine-tuning performance of \modelgene and \modelrank across different model sizes on the held-out donors task. We observe that both model variants improve in performance with the number of parameters, but \modelgene performs substantially better across all parameter sizes. Moreover, while \modelgene 160M and 400M outperform a task-specific XGBoost classifier trained on log-
transformed, filtered gene expression count data \modelrank does not.  

We also evaluated the impact of adding supervision through biological ontologies as described in \cref{sec:bioannot}. To this end, we compared the performance of \modelgene 70M and \modelrank 70M pre-trained with and without explicit supervision. On the downstream held-out donors task, \cref{fig:downstream_and_ontologies} summarizes our results. We find that both \model variants benefit from supervision; models pre-trained with biological ontologies result in better downstream performance. Moreover, we found that adding supervision led to more stable pre-training with fewer loss divergences and had no adverse impact on gene-modeling abilities of the model, as illustrated in \ref{tab:classification_comparison}. 

Having observed \modelgene to generally outperform \modelrank on the held-out donors task, we only experimented with \modelgene on the remaining tasks.

\subsection{\model models outperform existing foundation models on held-out donors task}
\cref{tab:donors_evaluation_results} summarizes the performance of different models on the held-out donors task. We observe that in comparison to competing foundation models, \modelgene generalizes better across donor variability. Upon finetuning, \modelgene 400M achieved an accuracy of 0.72, outperforming \nicheformer, the best-performing non-\model model, by 8.0\% (0.72 vs. 0.64) and \geneformer, the worst-performing model, by 45.8\% (0.73 vs. 0.39). Similarly, \modelgene improves the $\text{F}_1$ score over $\nicheformer$ by $17.6\%$ ($0.68$ vs. $0.56$) and a $67.6$\% improvement over $\geneformer$ ($0.68$ vs. $0.22$). These results highlight that \modelgene effectively generalizes across donors, an important use-case in practice.

\subsection{\model offers modest improvements over existing foundation models on held-out diseases task}

\begin{figure*}[t]
    \begin{center}
        \includegraphics[width=1\linewidth]{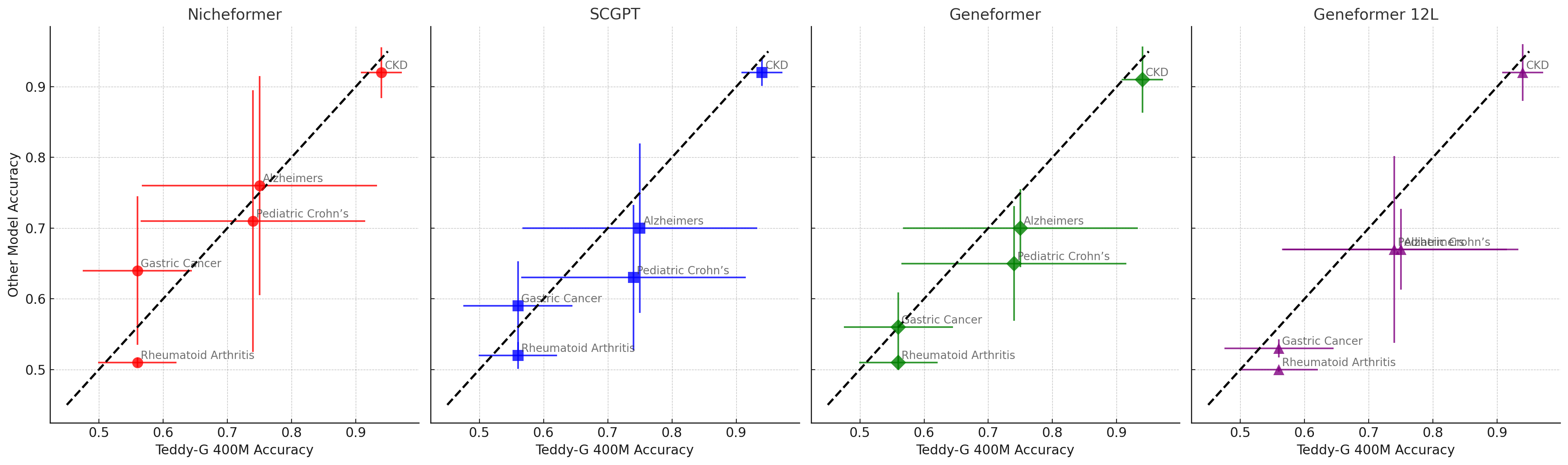}
    \end{center}
    \caption{\textbf{Performance of existing foundation models compared with \modelgene} \modelgene consistently outperforms other foundation models. The x-axis plots fine-tuning accuracy achieved \modelgene 400M. The y-axis is the accuracy achieved by other foundation models. Points below the diagonal indicate \modelgene 400M achieves higher accuracy.}
    \label{fig:diverse_diseases_downstream}
\end{figure*}

The performance of \modelgene 400M and other foundational models on the held-out diseases dataset is summarized in \cref{tab:diseases_evaluation_results} and \cref{fig:diverse_diseases_downstream}. 
We found that across the five binary classification sub-tasks \modelgene improves on all five when compared to \geneformer 12L, on all but one task compared to the six layered \geneformer and \scGPT. \modelgene's performance was within noise of \nicheformer, a model trained on similar volume of data as \model and an order of magnitude larger than \scGPT and \geneformer. This suggests that data volume is important for performing well on diseases not encountered during training. Scaling to larger volumes of data may lead to improved performance.    

\subsection{\model improves over task-specific machine learning approaches}
\begin{table}[t]
    \centering
    \caption{Logistic regression and \modelgene 400M on held-out diseases task.}
    \resizebox{0.5\textwidth}{!}{
    \begin{tabular}{l c c c}
        \toprule
        & \textbf{Logistic Regression} & \textbf{Linear probed \modelgene 400M} & \textbf{Finetuned \modelgene 400M} \\
        \midrule
        Chronic Kidney Disease & 0.90$\pm 0.03$ & 0.95$\pm 0.02$ & 0.94$\pm 0.03$ \\
        Alzheimer's Disease & 0.64$\pm 0.08$ & 0.86$\pm 0.07$ & 0.75$\pm 0.18$ \\
        Gastric Cancer & 0.49$\pm 0.04$ & 0.70$\pm 0.09$ & 0.56$\pm 0.09$ \\
        Rheumatoid Arthritis & 0.47$\pm 0.03$ & 0.93$\pm 0.10$ & 0.56$\pm 0.06$ \\
        Pediatric Crohn's Disease& 0.63$\pm 0.20$ & 0.72$\pm 0.16$ & 0.74$\pm 0.18$ \\
        \bottomrule
    \end{tabular}}
    \label{tab:lr_v_teddy_results}
\end{table}

We also benchmarked \model against a diverse set of task-specific machine learning methods---logistic regression, linear SVM, XGBoost, and lightGBM. We trained each of these methods to predict disease labels given log-transformed, filtered count data. We performed feature selection using the \texttt{highly\_variable\_genes} function from the \textsc{scanpy v1.10.1} library with default parameters. This method identifies genes exhibiting high variability across cells by calculating the normalized dispersion and selecting the top genes based on a predefined threshold. 
We performed the \texttt{highly\_variable\_genes} analysis for each cross-validation training fold and filtered the training and testing folds to only retain the identified genes.

In \cref{tab:lr_v_teddy_results} we highlight comparisons of fine-tuned \modelgene 400M and linearly (with logistic regression) probed \modelgene 400M against logistic regression. We note that the \model variants typically outperform logistic regression with linear-probing performing the best and often by a substantial margin. Additional results are detailed in \cref{tab:evaluation_results_ML}. Comparing these with \cref{fig:diverse_diseases_downstream} and \cref{tab:diseases_evaluation_results} we found that \modelgene 400M and \nicheformer~consistently improve over task-specific approaches. 

We conducted an additional experiment to evaluate whether the representations learned by \modelgene are useful for disease classification. To do this, we replaced the hand-engineered features used by the task-specific methods with pre-trained embeddings from \modelgene 400M. We constructed cell-level embeddings by averaging over output gene embeddings produced by \modelgene 400M. We note that these embeddings are zero-shot, they were not fine-tuned on the held-out disease datasets. We then trained the task specific methods as before. The results are summarized in \cref{fig:diverse_diseases_tml}, with numbers available in \cref{tab:evaluation_results_ML}. Observe that across all methods and all diseases replacing hand-engineered features with zero-shot embeddings leads to substantially better performance. These results suggest that embeddings better capture disease-relevant information than hand-engineered features based on log-transformed transcript counts.

\subsection{Zero-shot \model embeddings produce coherent clusters}
Next, to probe the zero-shot embeddings produced by \modelgene we looked at the embeddings produced by \modelgene 400M on data from~\cite{thomas2024longitudinal} that contains human gut single cell transcriptomes from donors with Ulcerative Colitis (UC), Crohn's Disease (CD), and those not afflicted by either disease. Importantly for our purposes, \citep{thomas2024longitudinal} provided detailed cell-type annotations at varying levels of granularity, allowing us to benchmark our embeddings quantitatively. We used a version of the data reprocessed using Cell-Ranger v7.1.0. After filtering out cells with greater than  $10\%$ mitochondrial proportion and those with no author-provided labels, we ended up with 536,207 cells with cell-type annotations. We benchmarked \modelgene 400M against a recently proposed single-cell foundation model  -- Cell2Sentence~\cite{rizvi2025scaling}. For this comparison, we selected the Cell2Sentence-410M model, which is of similar scale (410 million parameters) and fine-tuned for cell type prediction tasks. 

In \cref{fig:cell_type_umaps}, we plot two-dimensional embeddings from both models and present quantitative clustering metrics computed using the scib~\cite{luecken2022benchmarking} package in \cref{tab:clustering_metrics}. We found \modelgene 400M to produce largely coherent clusters and outperform Cell2Sentence 410M at both the coarse (low) and intermediate cell-type annotation resolutions. In particular, we found \modelgene 400M embeddings capable of distinguishing subtle cell-state differences, for instance, between CD\_4 and CD\_8 cell-states that Cell2Sentence fails to separate. Additional details about this experiment are in \cref{app:zero_shot_details}.

\subsection{In-silico single-gene perturbations with \model align with experimental evidence}
\label{sec:gata4}

To probe whether \modelgene encodes \emph{causal} gene–network structure, beyond merely capturing co-expression statistics, we used the single-gene knockout benchmark introduced as an evaluation for \geneformer ~\citep{theodoris2023transfer}. We evaluated the models with an \emph{in\,silico} single-gene perturbation that mimics a loss-of-function experiment for the cardiac transcription factor GATA4. The assay is conceptually analogous to an in-vitro CRISPR screen, but is performed entirely in-silico on the pretrained embedding space.
Crucially, we did not update any model parameters for this experiment, so these results reflect \emph{zero-shot} mechanistic knowledge acquired by the model during pre-training and underscores the potential of large pretrained models for \textit{in-silico} target discovery.

First, we isolated two-hundred-thirty-six fetal-cardiomyocyte transcriptomes from the heart cell atlas~\citep{knight2022single} in which GATA4 is expressed, so that the knockout is biologically relevant. We then created a perturbed representation of these cells by setting the GATA4 expression to zero. We then embedded the before and after perturbation cells using \modelgene (and other competing models).
For every gene \(g\) expressed in these cells we extracted its token-level representation before (\(\mathbf{v}^{\text{before}}_{g}\)) and after (\(\mathbf{v}^{\text{after}}_{g}\)) the perturbation and measured, $s_{g}\;=\;\cos\!\bigl(\mathbf{v}^{\text{before}}_{g},\,\mathbf{v}^{\text{after}}_{g}\bigr)$,
the cosine similarity between the two embeddings.  This yielded one similarity score per gene per cell.  We further partitioned genes into biologically annotated groups: \emph{direct targets} of GATA4, \emph{indirect targets}, and \emph{housekeeping genes}, using labels  provided in~\citep{theodoris2023transfer}.  For each cell, we averaged the similarity scores of the genes within each group to obtain average per-cell effects,
\(\bar{s}^{(i)}_{\text{direct}}\), \(\bar{s}^{(i)}_{\text{indirect}}\), \(\bar{s}^{(i)}_{\text{direct+indirect}}\) and \(\bar{s}^{(i)}_{\text{housekeeping}}\), with $i$ indexing the cell.  We then subjected the sets of similarities, $\{\bar{s}^{(i)}_{\text{direct}}\}_{i=1}^{236}, \{\bar{s}^{(i)}_{\text{indirect}}\}_{i=1}^{236}, 
\{\bar{s}^{(i)}_{\text{direct+indirect}}\}_{i=1}^{236},
\{\bar{s}^{(i)}_{\text{housekeeping}}\}_{i=1}^{236}$ 
to paired one-sided Wilcoxon signed-rank tests
whose alternative hypothesis matched the biological expectation
(e.g.\ $\bar s_{\text{housekeeping}}>\bar s_{\text{direct}}$ for the
\textit{HK-vs-Direct} comparison, and
$\bar s_{\text{housekeeping}}<\bar s_{\text{indirect}}$ for
\textit{Indirect-vs-Direct}). The four comparisons were (i) \textit{housekeeping} vs.~\textit{direct}, (ii) \textit{housekeeping} vs.~\textit{indirect}, (iii) \textit{housekeeping} vs.~\textit{direct+indirect}, and (iv) \textit{indirect} vs.~\textit{direct} and the resulting $p$-values are in \Cref{tab:gata4_wilcoxon}.
\begin{table}[htbp]
  \centering
  \resizebox{\linewidth}{!}{
  \begin{tabular}{lcccc}
    \toprule
      Model &
      $p_{\text{HK}-\text{Dir}}$ &
      $p_{\text{HK}-\text{Indir}}$ &
      $p_{\text{HK}-\text{Dir+Indir}}$ &
      $p_{\text{Indir}-\text{Dir}}$ \\ \midrule
      TEDDY-G 400M &
      \textbf{$9.45\times10^{-25}$} &
      \textbf{$6.47\times10^{-22}$} &
      \textbf{$1.66\times10^{-23}$} &
      \textbf{$1.27\times10^{-20}$} \\

      Geneformer 12-layer &
      \textbf{$1.43\times10^{-5}$} &
      \textbf{$7.72\times10^{-6}$} &
      \textbf{$9.26\times10^{-7}$} &
      \textbf{$0.96$} \\

      Geneformer (base) &
      \textbf{$5.01\times10^{-6}$} &
      \textbf{$4.59\times10^{-6}$} &
      \textbf{$2.22\times10^{-6}$} &
      \textbf{$0.98$} \\

      Nicheformer &
      $0.72$ &
      \textbf{$2.62\times10^{-6}$} &
      \textbf{$9.21\times10^{-4}$} &
      \textbf{$1.00$} \\

      scGPT &
      $0.75$ &
      $0.20$ &
      $0.32$ &
      \textbf{$0.99$} \\
    \bottomrule
  \end{tabular}}
    \caption{Paired Wilcoxon signed-rank $p$-values for the GATA4 knockout assay
           ($n=236$ cells; four tests per model).}
    \label{tab:gata4_wilcoxon}
\end{table}
Across all four paired comparisons, the \modelgene 400M model yields small \(p\)-values, showing that its embedding space separates gene sets exactly as GATA4 biology predicts: housekeeping genes change the least, direct targets change the most, indirect
targets lie in between, and the indirect–vs.–direct contrast is strongly negative.  Both \geneformer~ variants reproduce a similar qualitative ordering, but fail to distinguish between direct and indirect effects. 
\nicheformer
captures only downstream effects: it distinguishes indirect targets from housekeeping genes yet fails to separate direct targets, suggesting it encodes transcriptomic consequences rather than primary transcription factor–target relationships. \scGPT~ shows no significant housekeeping–target separation at all, indicating that its latent space is largely insensitive to this perturbation.

\noindent
Taken together, these results underscore two points:  
(i) increasing model scale and using biologically informed objectives markedly enhance zero-shot mechanistic knowledge, and  (ii) \modelgene can go beyond co-expression in the case of GATA4, and arrange genes in a manner that mirrors their causal proximity to the perturbed transcription factor.


\section{Acknowledgements}
The authors would like to thank Marinka Zitnik (Harvard Medical School, Harvard Data Science Initiative, Kemper institute, Broad Institute) for her thoughtful comments on the paper and to Michael Brochu (BCG) for overall support and direction.  We also would like to thank Greg Hersch and many other colleagues that enabled the infrastructure for training \model: (Merck Central IT) Ron Kim, Asheesh Chhabra, Manoj Vig, Amany Basily, Venki Balakrishnan, Mike Dickmann, Mohan Kumar, Rasti Matus, Kate Lipina, Amal Verghese, Karthik Palanis; (Merck Research Laboratories IT) Carol Rohl, Venkat Parakala, Antong Chen, Danny Bitton, Faisal Hoda, Atul Singal, Martin Radocha, Jan Lubojacky.
And thank you to our partners at Databricks: Pedro Portilla, Srijit Nair, Tim Lortz, and Abraam Samuel. 
\section*{Impact Statement}
This paper presents work whose goal is to advance the use of machine learning in molecular biology. There are many potential societal consequences of our work, none which we feel must be specifically highlighted here.

\bibliography{bibliography}
\bibliographystyle{plainnat}

\newpage
\appendix
\onecolumn
\section*{Appendix}
\setcounter{table}{0}
\renewcommand{\thetable}{A\arabic{table}}
\renewcommand{\thefigure}{A\arabic{figure}}
\setcounter{figure}{0} 
\section {Discussion}
\label{sec:app_limit}
We have introduced the \model family foundation models and developed a benchmark for probing the disease biology learned by these models. The \model family, trained on a comprehensive 116M dataset taken from the \cxg corpus, consists of two variants, \modelgene trained with masked gene modeling and \modelrank trained with masked gene expression level modeling. Both come in 70M, 160M, and 400M parameter sizes, and our empirical validation found that larger models and data led to improved performance, though with diminishing returns. Deviating from previous work, the \model models also used large-scale biological annotations as supervisory signals while pre-training. Our experiments showed that the inductive biases introduced by such supervision substantially improve downstream performance on the held-out donors task and do not hurt performance on the held-out diseases task. 

The limited effectiveness of the current crop of transcriptomic foundation models, including \model, on the held-out diseases task, could be attributed to either the failure of the models to adequately generalize to data not seen during pre-training or to the fact that the task itself is plagued by noise from experimental and data collection artifacts. After all, we do not have healthy or diseased annotations for individual cells; instead, we rely on whether the cells were derived from healthy or diseased donors. Carefully disentangling these factors is important for future work. 

Other future directions include improvements in the data aspects, such as increasing the size of the pre-trained corpus, quantifying the diversity of the corpus, augmenting it with additional disease-specific data, and more thoroughly filtering it to remove data of poor quality, as well as complimentary modeling improvements -- priors that carefully incorporate known biology improve model performance. Augmenting the training data with large scale interventional data~\citep{Tahoe-100M} and uniformly processed observational data~\citep{scbase} that reduce batch effects are of particular interest.  Another direction includes evaluations on more complex biological tasks, like perturbation prediction and inference of gene regulatory networks (GRNs). Foundation models that can accurately predict expression changes after gene perturbations \emph{in-silico} would be powerful tools for reconstructing disease mechanisms by identifying key regulatory pathways. By modeling how specific gene knockouts or activations impact downstream gene expression, these models could help biologists pinpoint the drivers of dysregulated pathways in diseases and prioritize targets for therapeutic intervention.

\model models could help facilitate the reconstruction of the GRN in two different ways. First, we plan to introduce a new fine-tuning task in which the model predicts the probability that a regulatory edge exists between two genes, thus enabling the identification of novel interactions that are otherwise more difficult to extract from experimental data. Second, leveraging Perturb-seq fine-tuned models to use predicted expression profiles after gene knockouts as input for GRN inference algorithms offers a data-driven approach to constructing regulatory networks. This dual strategy could empower biologists to map regulatory interactions efficiently and on a large scale, thereby accelerating discoveries in complex regulatory networks and their roles in disease biology.
Both these tasks directly apply to drug target discovery and a better understanding of disease biology. Finally, capturing multi-omics information via multi-modal integration would be critical, considering that cellular regulation is multi-layered. Hence, we plan to extend \model beyond its current scRNA-seq modality. Notwithstanding, \model establishes a new state-of-the-art for foundation models for scRNA-seq data. 

\section{Modeling Choices and Details}
\label{sec:details}

\subsection{Architecture details and model sizes}
The \model models vary in the number of parameters from approximately 10 million parameters to approximately 400 million parameters. The exact parameter counts are detailed in \cref{tab:pretrain_corpus_model_params_transposed}. The 10M \model models contain three transformer blocks (layers) and use a token embedding dimension of 128. The 30M models use six transformer blocks and a embedding dimension of 256. The 70M models use 12 transformer blocks and an embedding dimension of 512. The 160M models use 12 transformer blocks and an embedding dimension of 768 and finally the 400M models use 24 transformer blocks and 1024 dimensional embeddings.
\begin{table*}[t]
    \centering
    \caption{Training data size and model size for \model models and existing foundation models.} 
    \resizebox{1\textwidth}{!}{
    \begin{tabular}{@{}lcccccccc@{}}
        \toprule
        & \textbf{Pretrain corpus size by number of cells} & \textbf{Model size by number of parameters} \\
        \midrule
        \textbf{\model-G 400M} & 116M & 394.2M \\
        \textbf{\model-X 400M} & 116M & 414.2M \\
        \textbf{\model-G 160M} & 116M & 154.0M \\
        \textbf{\model-X 160M} & 116M & 164.8M \\
        \textbf{\model-G 70M} & 116M & 71.2M \\
        \textbf{\model-X 70M} & 116M & 72.0M \\
        \textbf{\model-G 30M} & 116M & 26.2M \\
        \textbf{\model-X 30M} & 116M & 26.1M \\
        \textbf{\model-G 10M} & 116M & 11.9M \\
        \textbf{\model-X 10M} & 116M & 11.8M \\
        \textbf{Nicheformer} & 110M & 46.8M \\
        \textbf{SCGPT} & 33M & 51.3M \\
        \textbf{Geneformer} & 30M & 10.3M \\
        \textbf{Geneformer 12L} & 30M & 39.6M \\
        \textbf{Cell2Sentence} & 57M & 410M \\
        \bottomrule
    \end{tabular}}
    \label{tab:pretrain_corpus_model_params_transposed}
\end{table*}

\subsection{Annotation labels} During pre-training, we balance the classes used for ontology classification. To achieve this, we compute the number of cells with each label over the entire train set, and during training, we randomly prompt the model with classification tokens with sampling probabilities designed to balance the probability of each label. 

In particular, we balance disease classification so that the cells with \texttt{<disease>} tokens have a 50\% probability of being normal or diseased. 

\subsection{Other design explorations for the \model family}\label{subsec:other_design}
To guide our model development efforts, we also  explore the effect of different approaches to data pre-processing, the effect of sequence length, and the effect of training on multi-species data. 

We implemented the median scaling method introduced in \cite{theodoris2023transfer}. We compute the median non-zero expression value of each gene after scaling each cell to 10,000 gene expression counts. We then scale each gene expression by this median value, so that each gene has a  median expression value of $1$ over the entire training dataset. We trained models with and without median-scaling on a subsampled version of our pre-training dataset comprising of four million (4M) cells. The model trained on unscaled data achieved a 60\% improvement in training loss, clearly indicating the importance of median scaling. All our models are trained with  median scaling.

We also compared the effect of pre-training with different sequence lengths. 

We trained two \modelgene models (without biological annotations) on our 4M-cell dataset with sequence lengths of 2,048 and 1,500. After truncating the \emph{test set} for both models down to 1,500-long sequences, the 2,048-long model achieved a test loss of 2.84, whereas the 1,500-long model achieved a loss of 2.88. The loss achieved by the 2,048-long model on the full 2,048-long sequences was 2.63. This shows that single-cell foundation models benefit from pre-training on longer sequences and understand shorter sequences better. Therefore, we train all our models with 2,048 tokens. We leave it as future work to study the importance of sequence length, especially in relation with recent work on long-context large language models.

Our 116M pre-training dataset contains both human and mouse genes. We tokenize ENSMUS mouse genes separately from human genes. Preliminary experiments on the 4M-cell data comparing models trained on both human and mouse data show that they achieve similar test loss to a model trained exclusively on human data. A more large scale exploration of using multi-species data is needed to better understand the benefit of jointly training on multi-species data and is interesting future work, especially with the advent of large multi-species datasets~\citep{scbase}.

\section{Evaluation benchmark details}
\label{app:evals}
\begin{figure}[th]
\includegraphics[width=0.5\textwidth]{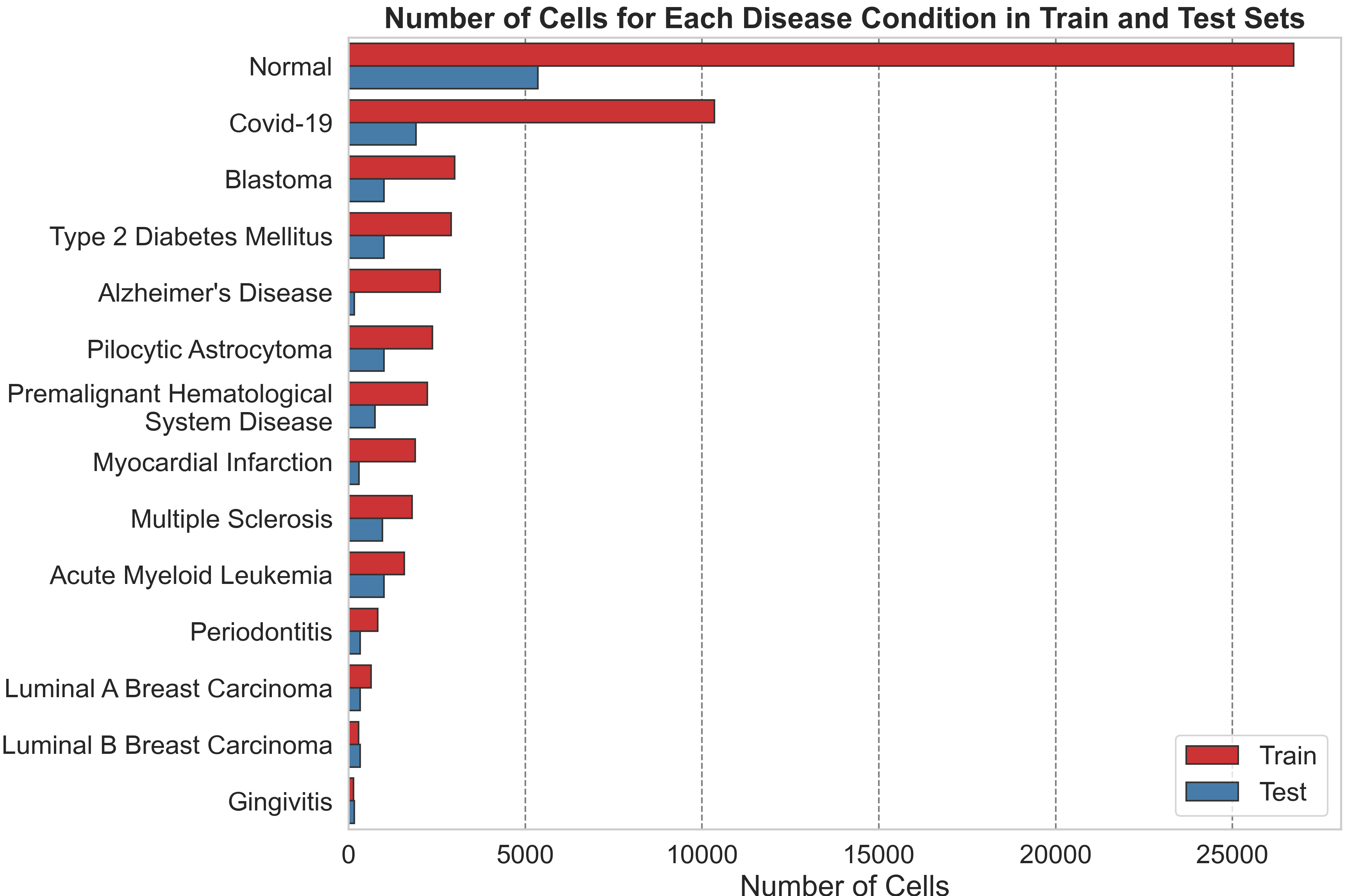}
\includegraphics[width=0.5\textwidth]{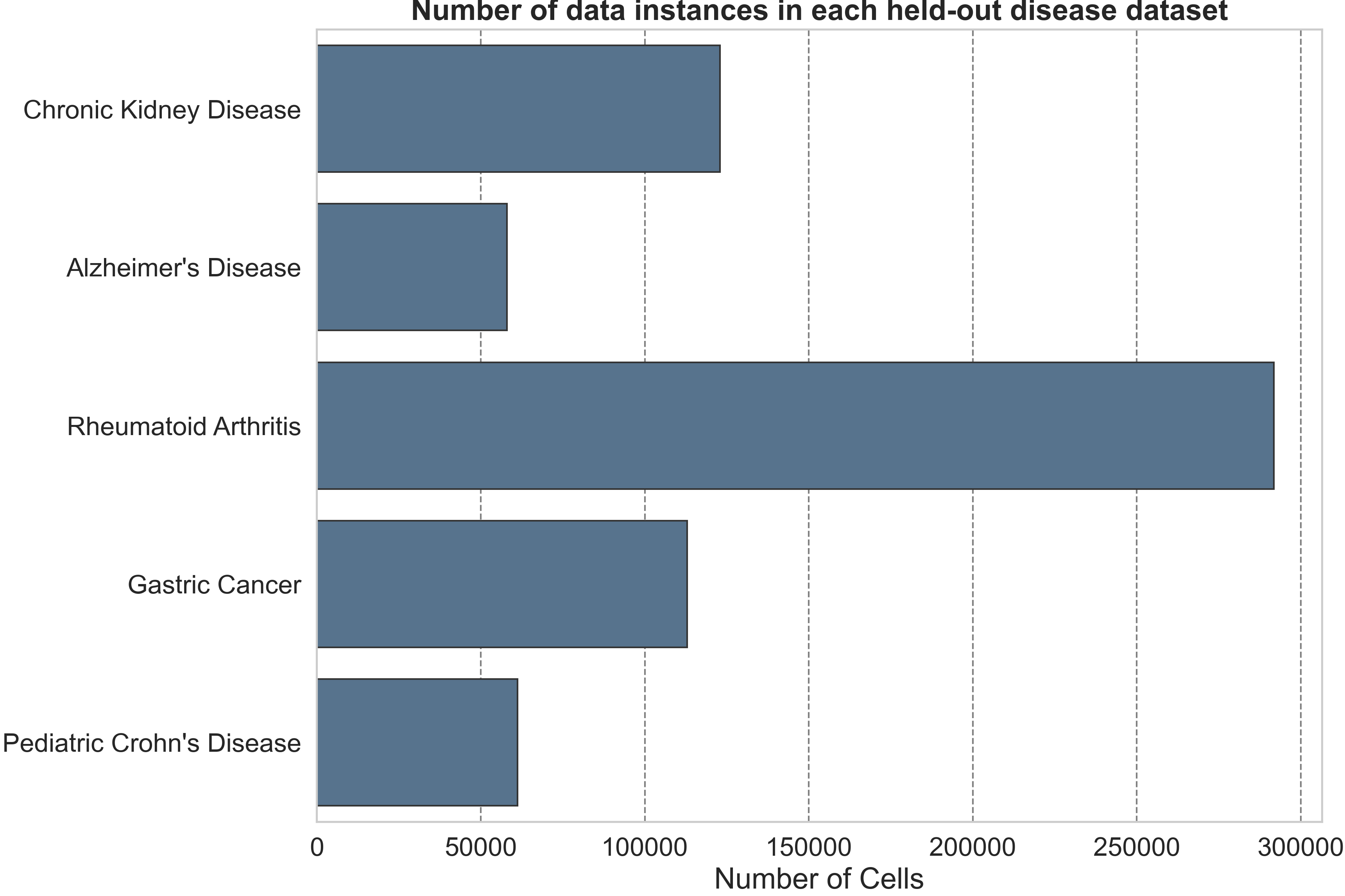}
\caption{\textbf{Evaluation benchmark statistics.} \emph{Left:} Distributions of disease conditions in train and test sets of the held-out donors dataset. \emph{Right:} Number of instances (cells) in the different held-out disease datasets. } 
\label{figure:app-labels}
\end{figure}
\subsection{Held-out donors dataset}
The training set comprises of cells from 524 donors.  
The test set, contains data from 82 donors who were excluded from the pre-training data. The task is a 14-way classification problem, \cref{figure:app-labels} illustrates the distribution of cells across the fourteen labels (thirteen disease conditions and healthy). As can be seen the proportion of labels vary widely across disease conditions. To account for this, we report class-weighted $\text{F}_1$ scores in addition to accuracy.  
\paragraph{Evaluation details} We used a single train/test split, with 70\% of the training set used for training and 30\% for validation. We used AdamW to fine-tune the models. We ran a parameter sweep over six learning rates (1E-3, 1E-4, 2E-4, 2E-5, 5E-4, 5E-5) and using pytorch default values for other optimizer parameters. The best learning rate was selected based on accuracy on the validation set. The final model was trained on the full training set and tested on the held-out set and repeated with three random seeds per model. 

\subsection{Held-out diseases dataset}
We use five datasets sourced from CellXGene for binary disease classification of cells as healthy or diseased in the held-out diseases task. These 5 datasets, and any other datasets containing disease labels present in these, were explicitly excluded from the pre-training corpus. The CellXGene datasets were preprocessed before inclusion, with minimal modifications applied to maintain their original state for reproducibility, except for filtering of non-primary datasets. The datasets chosen for evaluation were processed and required to meet several criteria: (1) inclusion of data from at least five donors to allow for robust cross-validation splits by donor; (2) an equal balance of healthy and diseased cells to address the class imbalance issue of the original data; and (3) a minimum of 10,000 cells per dataset, to provide sufficient training signal. We evaluated the performance of the model for the binary disease classification task using k-fold accuracy. Since these datasets are balanced between healthy and diseased cell types we just report accuracy as a measure of performance.

The dataset includes transcriptomic data from five different disease conditions, each with a balanced representation of diseased and normal cells. The Chronic Kidney Disease (CKD) dataset comprises 123,982 kidney cells from 25 donors, spanning 26 distinct kidney-related cell types, including epithelial cells of the proximal tubule, loop of Henle, and collecting ducts. The Alzheimer’s Disease dataset consists of 57,010 cells from the prefrontal cortex, sourced from 11 donors, all classified as inhibitory interneurons. The Gastric Cancer dataset contains 122,922 stomach-derived cells from 9 donors, but all are labeled under an unknown classification. The Pediatric Crohn’s Disease dataset features 61,244 cells from 16 donors, representing 31 diverse immune and epithelial cell types, such as enterocytes, B cells, and intestinal crypt stem cells, collected from the ileum. Lastly, the Rheumatoid Arthritis (RA) dataset is the largest, with 291,140 immune cells derived from 25 donors, predominantly consisting of T cells (CD4+ and CD8+), monocytes, dendritic cells, and natural killer cells, all obtained from blood samples. Together, these datasets offer a comprehensive resource for studying disease-related transcriptomic variations across multiple organ systems.

\paragraph{Evaluation details} We performed 3-fold cross-validation, creating three train/test folds such that there is no overlap in donors across the train and test folds and the proportion of diseased and normal cells in each fold is 50\%.   Each training fold was further divided into train and validation subsplits for model selection. We used AdamW to fine-tune the mdoels. We ran a parameter sweep over six learning rates (1E-3, 1E-4, 2E-4, 2E-5, 5E-4, 5E-5) on each subsplit and selected the best learning rate per model and disease based on accuracy on the validation sub-split. Finally we retrained the model on the full training fold with the selected hyper-parameter and tested on the unseen test fold. Since the test folds are perfectly disease-balanced so we reported only classification accuracy.

\section{Experimental details}
In \cref{tab:evaluation_results_ML}, \cref{tab:diseases_evaluation_results}, \cref{tab:classification_comparison},  \cref{tab:classification_comparison}, \cref{tab:teddy_models_donors_f1_scores} we provide detailed numbers from our expriments described in \cref{sec:exp} and \cref{app:evals}.
\begin{figure*}[t]
    \begin{center}
        \includegraphics[width=1\linewidth]{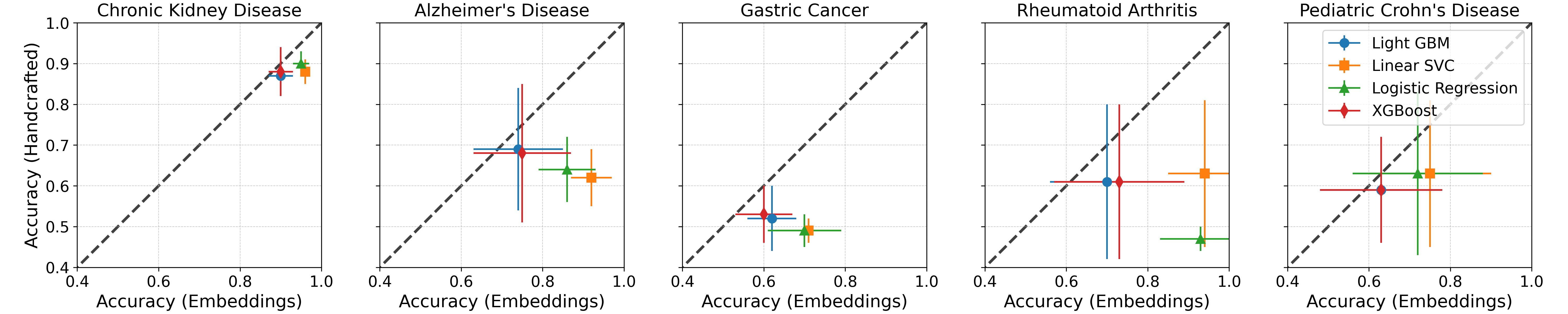}
    \end{center}
    \caption{\textbf{Performance of task-specific machine learning methods using handcrafted features and \modelgene 400M embeddings} Accuracy increases with use of \modelgene embeddings. The x-axis plots accuracy achieved by different methods using \modelgene 400M embeddings as features. The y-axis is the accuracy achieved when using hand engineered features. Points below the diagonal indicate that embedding features achieve higher accuracy.}
    \label{fig:diverse_diseases_tml}
\end{figure*}
\begin{table}[ht]
    \centering
    \caption{Accuracy of task-specific methods on held-out diseases task.}
    \resizebox{1\textwidth}{!}{
    \begin{tabular}{l c c c c}
        \toprule
        & \textbf{LightGBM} & \textbf{Linear SVC} & \textbf{Logistic Regression} & \textbf{XGBoost Classifier}  \\
        \midrule
        \multicolumn{4}{l}{\textbf{Held-out diseases - Handcrafted features}} \\
        \midrule
        CKD & 0.87$\pm 0.05$ & 0.88$\pm 0.03$ & 0.90$\pm 0.03$ & 0.88$\pm 0.06$ \\
        Alzheimers & 0.69$\pm 0.15$ & 0.62$\pm 0.07$ & 0.64$\pm 0.08$ & 0.68$\pm 0.17$ \\
        Gastric Cancer & 0.52$\pm 0.08$ & 0.49$\pm 0.03$ & 0.49$\pm 0.04$ & 0.53$\pm 0.07$ \\
        Rheumatoid Arthritis & 0.61$\pm 0.19$ & 0.63$\pm 0.18$ & 0.47$\pm 0.03$ & 0.61$\pm 0.19$  \\
        Pediatric Crohn's & 0.59$\pm 0.13$ & 0.63$\pm 0.18$ & 0.63$\pm 0.20$ & 0.59$\pm 0.13$ \\
        \midrule
        \multicolumn{4}{l}{\textbf{Held-out diseases - \modelgene embeddings as predictive features}} \\
        \midrule
        CKD & 0.90$\pm 0.03$ & 0.96$\pm 0.01$ & 0.95$\pm 0.02$ & 0.90$\pm 0.03$ \\
        Alzheimers & 0.74$\pm 0.11$ & 0.92$\pm 0.05$ & 0.86$\pm 0.07$ & 0.75$\pm 0.12$ \\
        Gastric Cancer & 0.62$\pm 0.06$ & 0.71$\pm 0.08$ & 0.70$\pm 0.09$ & 0.60$\pm 0.07$ \\
        Rheumatoid Arthritis & 0.70$\pm 0.14$ & 0.94$\pm 0.09$ & 0.93$\pm 0.10$ & 0.73$\pm 0.16$  \\
        Pediatric Crohn's & 0.63$\pm 0.15$ & 0.75$\pm 0.15$ & 0.72$\pm 0.16$ & 0.63$\pm 0.15$ \\
        \bottomrule
    \end{tabular}}
    \label{tab:evaluation_results_ML}
\end{table}
\begin{table}[ht]
    \centering
    \caption{Accuracy of different foundation models on held-out diseases task.}
    \resizebox{1\textwidth}{!}{
    \begin{tabular}{@{}lccccc@{}}
        \toprule
        & \textbf{\model-G 400M} & \textbf{Nicheformer} & \textbf{SCGPT} & \textbf{Geneformer} & \textbf{Geneformer 12L} \\
        \midrule
        \multicolumn{5}{l}{\textbf{Held-out diseases - Fine-tuning}} \\
        \midrule
        CKD & {0.94$\pm 0.03$ } & {0.92$\pm 0.04$} & {0.92$\pm 0.02$} & {0.91$\pm 0.05$} & {0.92$\pm 0.04$} \\
        Alzheimers &  {0.75$\pm 0.18$} & {0.76$\pm 0.16$} & {0.70$\pm 0.12$} & {0.70$\pm 0.06$} & {0.67$\pm 0.06$}\\
        Gastric Cancer & 0.56$\pm 0.09$ & {0.64$\pm 0.11$} & {0.59$\pm 0.06$} & {0.56$\pm 0.05$} & {0.53$\pm 0.01$} \\
        Rheumatoid Arthritis &  {0.56$\pm 0.06$} & {0.51$\pm 0.01$} & {0.52$\pm 0.02$} & {0.51$\pm 0.01$} & {0.50$\pm 0.00$} \\
        Pediatric Crohn's & {0.74 $\pm 0.18$} & {0.71$\pm 0.19$} & {0.63$\pm 0.10$} & 0.65$\pm 0.08$ & 0.67$\pm 0.13$ \\
        \bottomrule
    \end{tabular}}
    \label{tab:diseases_evaluation_results}
\end{table}
\begin{table}[ht]
\caption{Adding the ontology classification training objective to the gene modeling objective during pre-training does not affect gene-modeling loss, both for \modelgene and \modelrank.}
\centering
\begin{tabular}{lcccc}
     & Gene        \\
     & modeling loss \\
     \midrule
   \model-G                        & 2.48 &\\
   ~~~~~ - ontology classification (\model-G 70M)  & 2.48 & \\
   \model-X           & 0.123  \\
   ~~~~~ - ontology classification (\model-X 70M)    & 0.125 \\
\end{tabular}
\label{tab:classification_comparison}
\end{table}
\begin{table}[t]
    \centering
    \caption{Performance on held-out donors as a function of model size. Fine-tuning $F_1$ scores improve for both TEDDY-G and TEDDY-X with increasing size.}
    \resizebox{0.5\textwidth}{!}{
    \begin{tabular}{@{}lcc@{}}
        \toprule
        & \textbf{\modelgene} & \textbf{\modelrank} \\
        \midrule
        70M parameters & $0.65 \pm 0.02$ &  $0.52 \pm 0.00$ \\
       160M parameters & $0.70 \pm 0.03$ & $0.53 \pm 0.00$ \\ 
        400M parameters & $0.68 \pm 0.06$ & $0.55 \pm 0.03$ \\ 
        \bottomrule
    \end{tabular}}
    \label{tab:teddy_models_donors_f1_scores}
\end{table}

\subsection{Zero-shot embeddings evaluation}
\label{app:zero_shot_details}
\begin{table*}[t]
\centering
\caption{\modelgene 400M offers better zero-shot clustering performance.}
\begin{tabular}{@{}lcccccccc@{}}
\toprule
                            & \multicolumn{4}{c}{\textbf{cell sub-clustering (low)}}            & \multicolumn{4}{c}{\textbf{cell sub-clustering (intermediate)}}            \\ \midrule
                            & \textbf{NMI} & \textbf{ARI} & \textbf{ASW} & \textbf{Average BIO} & \textbf{NMI} & \textbf{ARI} & \textbf{ASW} & \textbf{Average BIO} \\ \midrule
\textbf{Cell2Sentence 410M} & 0.7817       & 0.6385       & 0.5632       & 0.6611               & 0.7969       & 0.6768       & 0.5456       & 0.6731               \\
\textbf{Teddy-G 400M}       & 0.8382       & 0.7154       & 0.6022       & 0.7186               & 0.8785       & 0.7929       & 0.5971       & 0.7562               \\ \bottomrule
\end{tabular}
\label{tab:clustering_metrics}
\end{table*}

\begin{figure*}[t]
    \begin{center}
        \includegraphics[width=1\linewidth]{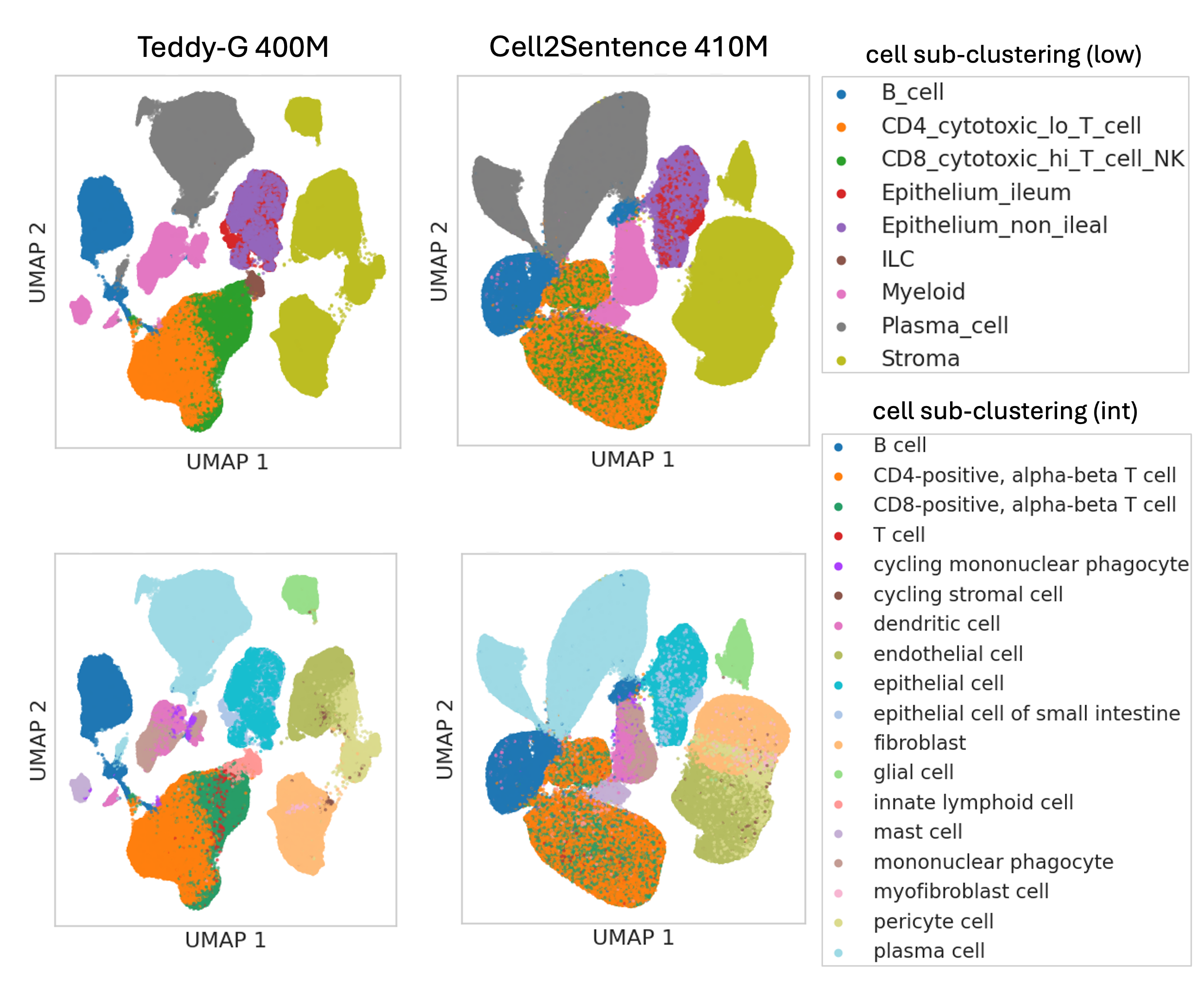}
    \end{center}
    \caption{\textbf{Zero-shot embeddings.} Two dimensional umaps of zero-shot embeddings from \modelgene 400M (left) and Cell2Sentence 410M (right). The top row labels the umaps with coarse-grained cell labels while the bottom row uses finer-grained labels.}
    \label{fig:cell_type_umaps}
\end{figure*}
We evaluated the clustering performance using four key metrics from scib~\cite{luecken2022benchmarking} package in \cref{tab:clustering_metrics}: Normalized Mutual Information (NMI), which measures the shared information between clustering results and true cell labels (0 to 1, higher is better); Adjusted Rand Index (ARI), which assesses similarity while adjusting for chance (-1 to 1, higher is better). NMI and ARI are computed based on Louvain clusters generated from the embedding space; Average Silhouette Width (ASW), which indicates cluster separation by computing within-cluster and between-cluster distances, and dividing this by the larger of the two values (0 to 1, higher is better); and Average Bio (AvgBIO), the arithmetic mean of ASW, NMI, and ARI. We used resolution optimized Leiden clustering using the \texttt{cluster\_optimal\_resolution} function with default settings provided by scib. 

In addition, we generate UMAPs using the following parameters in \cref{fig:cell_type_umaps}: we first perform Principal Component Analysis (PCA) on the dataset, setting the number of components to 50 with the ``svd\_solver'' set to ``arpack'' and a random seed of 42. Next, we compute the neighborhood graph with 15 neighbors, also using the same random seed. Finally, we apply UMAP to visualize the data, with the same seed.

\end{document}